%% file: blinded_main.tex
\documentclass[sn-nature]{sn-jnl}

\usepackage{amsmath,amssymb,amsfonts}%
\usepackage{amsthm}%
\usepackage{mathrsfs}%
\usepackage[title]{appendix}%
\usepackage{textcomp}%
\usepackage{manyfoot}%
\usepackage{booktabs}%
\usepackage{listings}%

\usepackage{natbib}
\usepackage{todonotes}

\usepackage{multirow} 
\usepackage{multicol} 

\usepackage[cp1252]{inputenc}
\usepackage{hyperref}

\usepackage{times}
\usepackage{graphicx}
\usepackage{epstopdf}

\usepackage{hhline}
\usepackage{latexsym}
\usepackage{xcolor}
\usepackage{lscape}

\usepackage{algorithm}
\usepackage{comment}

\usepackage{caption}
\usepackage[noend]{algorithmic}

\newcommand{\pezzonuovo}[1]{#1}

\newcommand{\tuple}[1]{\langle#1\rangle}
\graphicspath{{./exp/}}

\def\hb{\hbox to 10.7 cm{}}

\def\trace{\Phi}

\def\ma{\textsc{ma}}
\def\mb{\textsc{mb}}

\def\ecurr{\mbox{$e_{\mbox{\scriptsize{\emph{curr}}}}$}}

\def\U{\mbox{$\mathcal{U}$}}

\def\A{\mbox{$\mathcal{A}$}}
\def\M{\mbox{$\mathcal{M}$}}
\def\E{\mbox{$\mathcal{E}$}}

\newtheorem{example}{Example}
\newtheorem{definition}{Definition}

\def\accT{\mbox{\emph{Acc}$_{T}$}}
\def\accR{\mbox{\emph{Acc}$_{R}$}}
\def\accTA{\mbox{\emph{Acc}$_{T+A}$}}
\def\accTR{\mbox{\emph{Acc}$_{T+R}$}}
\def\timeT{\mbox{\emph{Time}$_{T}$}}
\def\timeTA{\mbox{\emph{Time}$_{T+A}$}}
\def\timeTR{\mbox{\emph{Time}$_{T+R}$}}

\usepackage{xargs}

\begin{document}

\title[]{Combining Abstract Argumentation and Machine Learning for Efficiently Analyzing Low-Level Process Event Streams}

%\author{AUTHOR BLINDED}

\author[1]{\fnm{Bettina} \sur{Fazzinga}}\email{bettina.fazzinga@unical.it}

\author[2]{\fnm{Sergio} \sur{Flesca}}\email{flesca@dimes.unical.it}

\author[2]{\fnm{Filippo} \sur{Furfaro}}\email{furfaro@dimes.unical.it}

\author*[3]{\fnm{Luigi} \sur{Pontieri}}\email{luigi.pontieri@icar.cnr.it}

\author[3]{\fnm{Francesco} \sur{Scala}}\email{francesco.scala@icar.cnr.it}

\affil*[1]{\orgdiv{DICES}, \orgname{University of Calabria}, 
\orgaddress{\street{Via Bucci}, \city{Rende (CS)}, \postcode{87036}, \country{Italy}}}

\affil[2]{\orgdiv{DIMES}, \orgname{University of Calabria}, 
\orgaddress{\street{Via Bucci}, \city{Rende (CS)}, \postcode{87036}, \country{Italy}}}

\affil[3]{\orgdiv{ICAR}, \orgname{CNR}, 
\orgaddress{\street{Via Bucci}, \city{Rende (CS)}, \postcode{87036}, \country{Italy}}}


\abstract{
Monitoring and analyzing process traces is a critical task for modern companies and organizations. In scenarios where there is a gap between trace events and reference business activities, this entails an \textit{interpretation problem}, amounting to translating each event of any ongoing trace into the corresponding step of the activity instance. Building on a recent approach that frames the \textit{interpretation problem} as an acceptance problem within an \textit{Abstract Argumentation Framework} (AAF), one can elegantly analyze plausible event interpretations (possibly in an aggregated form), as well as offer explanations for those that conflict with prior process knowledge. 
Since, in settings where event-to-activity mapping is highly uncertain (or simply under-specified) this reasoning-based approach may yield lowly-informative results and heavy computation, one can think of discovering a sequence-tagging model, trained to suggest highly-probable candidate event interpretations in a context-aware way. However, training such a model optimally may require using a large amount of manually-annotated example traces.
%Considering the need of dealing with limited amounts of activity-labeled trace developing solutions enabling environmental and societal sustainability (with reduced labor/computational costs and carbon footprint), 
We then propose a data-efficient neuro-symbolic approach to the problem, where the candidate interpretations returned by the example-driven sequence tagger is refined by the AAF-based reasoner. This allows us to also leverage prior knowledge to compensate for the scarcity of example data, as confirmed by experimental results. 
% This property is also beneficial in settings where data annotation and model optimization costs are subject to stringent constraints.
}

\keywords{Business Process Intelligence, Log Abstraction, Abstract Argumentation, Machine Learning}



\maketitle

\begin{comment}
    
\subsection*{Declarations}
\noindent \textbf{Funding} The authors received no specific funding for this study.\\
\textbf{Conflict of interest/Competing interests} The authors declare no conflict of interest.\\
\noindent 
\textbf{Ethics approval} Not Applicable.\\ 
\noindent 
\textbf{Data availability}
Data contain private information that cannot be disclosed.\\
\noindent 
\textbf{Authors' contribution}
All authors have contributed equally to this work.\\
\noindent 
\textbf{Acknowledgement}
This work was partly supported by project FAIR - Future AI Research - Spoke 9 (Directorial Decree no. 1243, August 2nd, 2022; PE 0000013; CUP B53C22003630006), under the NRRP (National Recovery and Resilience Plan) MUR program (Mission 4, Component 2 Investment 1.3) funded by the European Union - NextGenerationEU.

\thispagestyle{empty}

\newpage
\end{comment}

\section{Introduction}

Process Mining methods help better understand and enact business processes by supporting various analysis tasks over process \emph{traces}, i.e. sequences of execution \emph{events}, e.g. process-model discovery, 
log-to-model conformance analysis, runtime detection/prediction/recommendation.

The potential of these methods to turn log data into value is enormous in emerging application scenarios where complex and flexible operational/business processes have become easier to automate, monitor, analyze and improve, thanks to the widespread availability of activity tracking, IoT devices, and data gathering/storage/processing technology. Prominent scenarios of this kind range from Manufacturing (e.g., flexible production processes in Smart Factories), Healthcare (e.g., personalized care-flow/assisted-living processes), up to new-generation Logistics and Transportation. 

However, process mining methods usually assume each event to map to a well-understood ``high-level'' activity, which is not always the case in practice. 
In particular, when the activities are performed 
in a lowly-structured way as in the application contexts mentioned above, trace events just represent low-level actions~\cite{VanZelst2020} with no clear reference/mapping to the activities.

An example of this kind is presented below, for a toy care-flow, the traces of which consist of events describing exams and checks performed by medical staff. 

\medskip

\begin{figure}[!t]
	\includegraphics[width=0.75\textwidth]{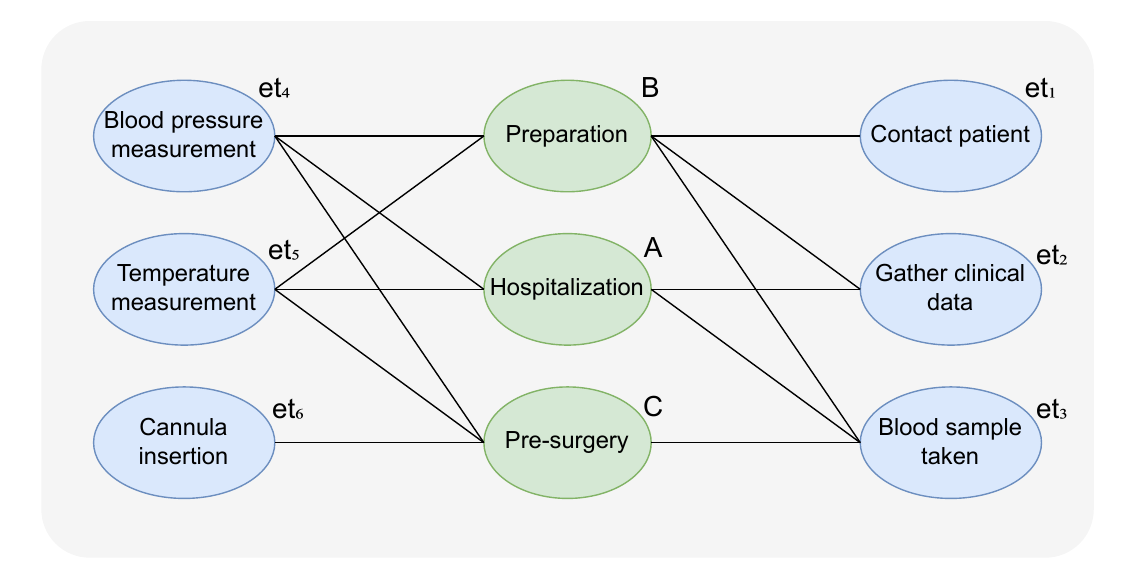}
	\centering
	\caption{Graphical representation of the mapping between the \underline{event types} and \underline{activities} of Example \ref{ex:example}.
    Green nodes represent the activities, blue nodes represent the event types, and \pezzonuovo{each edge between an event type, say $et$, and an activity, say $a$, means that an event of type $et$ (i.e. an instance of that event type) can be generated by an instance of activity $a$. The label and description et of each event type are reported by and within, respectively, the corresponding node. The same is done for the activity nodes.}
    }
	\label{fig:mapping}
\end{figure}

\begin{example}\label{ex:example}
Suppose that a trace  $\trace=[e_1,e_2,e_3,e_4,e_5,e_6]$ is given, where $e_1, e_2, e_3, \ldots, e_6$ refer to the \underline{event types} 
\pezzonuovo{$et_1$ (\emph{Contact patient}),
$et_2$ (\emph{Gather clinical data}), $et_3$ (\emph{Blood sample taken}), $et_4$ (\emph{Blood pressure measurement}), $et_5$ (\emph{Temperature measurement}), and $et_6$ (\emph{Cannula insertion}), respectively.
}
Assume that one wants to monitor process executions in terms of three\pezzonuovo{\underline{activities} (each of which represents a high-level action type)}: $B$ (\emph{preparation}), $A$ (\emph{hospitalization}), and $C$ (\emph{pre-surgery}), and that the following mapping, sketched in Figure \ref{fig:mapping}, relates these activities to the above-mentioned event types: 
\pezzonuovo{events of type $et_1$ can only be performed during (an instance of) activity $B$, events of type $et_2$ can be performed during (instances of) both activities $B$ and $A$, events of types $et_3,et_4$ and $et_5$ can be performed during any of the activities $A$, $B$ and $C$, while events of type $et_6$ can be performed only during (an instance of) activity $C$.}
To reconstruct/monitor patient histories, one needs to interpret such a trace in terms of activity executions. 
\pezzonuovo{However, due to the many-to-many nature of the  \underline{mapping}\footnote{\pezzonuovo{The term \emph{mapping} here refers to the relationship between two families of types, modeling process execution steps at different abstraction levels: event types (representing the types of low-level actions log events refer to) and activities (representing high-level process tasks). Following the mainstream terminology in Process Mining literature, the instances of an event type are called events. In fact, in our and other Process Mining settings, one is interested in analyzing the events recorded in a process log; since each of these events is required to refer to a single event type, it is natural to look at the former as an instance of that event type.}} between event types and activities,} even if trace $\trace$ describe a completed process instance, multiple interpretations may exist for $\trace$, like: 
$I_1 \equiv$ ``\emph{$e_1,e_2$ is the sequence of steps produced by an instance of $B$,
$e_3,e_4,e_5$ is the sequence of steps produced by an instance of $A$ and $e_6$ is the unique step of an instance of $C$}''; 
and $I_2 \equiv$ 
``\emph{$e_1,e_2$ is the sequence of steps produced by an instance of $B$, while
$e_3,e_4$ and $e_5,e_6$ are the step sequences produced by instances of $A$ and of $C$, respectively}''.
\hfill $\Box$
\end{example}
\medskip

This abstraction gap limits the effectiveness of process discovery methods and hinders the application of standard conformance-checking techniques. Nevertheless, enabling business-level monitoring of loosely structured processes is essential for assessing their quality and compliance with requirements, as well as for promptly detecting and mitigating undesired events such as failures or exceptions

\subsection{State of the art: event abstraction/interpretation solutions and their limitations}
\emph{Supervised log abstraction} solutions were proposed 
recently~\cite{VanZelst2020}, 
which try to \emph{interpret} log traces automatically as executions of given high-level process activities, 
leveraging prior information concerning some reference process activities, in terms of which the given log traces must be interpreted. 
Supervised event abstraction approaches can be classified in two broad families, named hereinafter 
\emph{example-driven} and \emph{knowledge-driven}, according to the kind of domain information exploited in 
the interpretation task (besides the mere activity labels), i.e.: 
(a) annotated example traces~\cite{tax2018human,abs-CRF}, for training some kind of machine learning model, or
(b) formally represented domain knowledge~\cite{paper,Baier-conf15,Baier2018,Baier-IS14,Baier-conf14,nostro-IS18,Ferreira14,abs-ActivityPattern,leonardi2017,abs-ROAD}, 
allowing for pruning out meaningless trace interpretations.
Domain knowledge is usually expressed via declarative behavioral constraints, such as DECLARE rules~\cite{declare} (very popular in the field of Process Mining), which correspond to unary and binary templates of Linear Temporal Logics formulae.
\pezzonuovo{
Most of these approaches try to match each (low-level) event to a single activity instance, even if its event type has multiple activities associated in the mapping.} In general, this leads to a loss of information on the plausible interpretations of the event and of the trace containing it, especially in run-time monitoring settings where each event has to analyzed as soon as it is produced: since the ``best'' interpretation of each event can be chosen only on the basis of the events preceding it, without having the possibility to look at the rest of the trace, event interpretation errors may arise and accumulate, possibly leading to suboptimal results or to the degenerate situation where no admissible interpretations exist for all the events of a trace from a certain step on.
Further, none of previous solutions in the field but in \cite{paper} was conceived to provide user with mechanisms for obtaining explanations on the interpretation/abstraction results returned.

\paragraph{An existing approach based on Abstract Argumentation}
In fact, the framework proposed in \cite{paper}  overcomes the two major limitations mentioned above by allowing the user to obtain answers and explanations for \textit{Interpretation Queries} over any step $e_{curr}$ of a low-level $\trace$ as soon as it is produced (or, possibly, later on), so supporting interactive explorative analyses of log data.  

In a nutshell, in this framework, interpretation queries can be used to ask whether $e_{curr}$ can be interpreted as a step of an instance of a given activity type $A$. 
Several query variants are supported by the framework, which specifically allow the user to state conditions on the internal activity step, % (e.g., is $e_{curr}$ the initial step of an instance of $A$?") 
and/or the ordinal number of the activity instance,
as well as for listing all valid interpretations of $e_{curr}$ satisfying the conditions required.
The computation of both answers to interpretation queries and explanation relies on encoding the underlying event-interpretation problem 
into an \emph{Abstract Argumentation Framework} (\emph{AAF}), which incorporates the given domain 
knowledge on both candidate event-activity mappings and activity-level behavioral constraints, and on solving specific instances 
of the acceptance problem over the AAF.
More details on the queries and computation aspects of the AAF-based approach proposed in \cite{paper} are briefly illustrated in Section~\ref{sec:background}, which also provide some simple application examples for the sake of readability.

Notably, the ability to reason about alternative interpretations of log data and their associated explanations can be extremely valuable in application contexts where multiple valid interpretations exist for an event or a trace. This is often due to the fact that the domain knowledge available is not enough to fully resolve interpretation ambiguities. In such contexts, users must be promptly provided with interpretable (possibly aggregated) representations of log events that are as complete and reliable as possible.
% In fact, the possibility of reasoning on alternative interpretations of log data and on associated explanations can turn very precious in application 
% settings where multiple valid interpretations do exist for an event/trace, just because domain knowledge is not enough to dissolve all interpretation 
% ambiguity, and where the users need to be provided timely with interpretable log event representations that are as complete and trustworthy as possible.

However, as observed in \cite{paper}, if the known behavioral constraints are loose, the result of an interpretation query may contain too many items 
to be presented as-is to the user. 
%To solve this information-load issue, mechanisms supporting ranked and/or top-k enumeration queries can be devised. 

\subsection{Proposed approach and contributions}

In this paper, we propose a novel approach that combines example-driven abstraction methods~\cite{tax2018human,abs-CRF} with the constraint-driven abstraction technique introduced in~\cite{paper}. This integration leads to two main contributions:

\begin{enumerate}
\item It enhances the method proposed in~\cite{paper} by 
leveraging data-driven knowledge to present the interpretations of an event in a ranked (the higher the interpretation probability the higher the rank) and possibly partial (only the top-k interpretations) way, %of an event , 
thereby addressing the information overload that may arise when constraints are too loose;
\item It improves the performance of example-driven abstraction methods~\cite{tax2018human,abs-CRF} by leveraging process model knowledge, which is particularly beneficial in scenarios where the available examples of traces do not suffice to learn a tight characterization of the process.
\end{enumerate}

As regards the first contribution, instead of just assigning global belief weights to candidate event-activity pairs as envisaged in \cite{paper}, we here propose to train a sub-symbolic 
Machine-Learning (ML) model over given activity-labeled log traces (as in previous example-driven abstraction methods~\cite{tax2018human,abs-CRF}) 
to learn to map any event $e$ of a trace $\trace$ to the activities most likely generated $e$ in a context-depended way (i.e., taking into account all the events stored 
in $\trace$, rather than the sole event $e$ itself). 

As for the second contribution, the accuracy of a sequence-tagging model is likely to depend on the quality and representativeness of the sample example traces 
(and associated activity labels) that were used to train it. 
Since annotating process traces with per-step activity labels is a costly task for human experts, in terms of both skills and time, it may well be the case 
that the training examples at hand are not enough to obtain a good-quality tagging model, especially if it is a powerful (but over-parametrized) neural 
network like the ones proposed in~\cite{tax2018human}. 
In addition, %owing to the increasing attention towards Green-aware AI applications~\cite{adadi2021}, 
the amount of log traces available for training  can be subject to data access/processing operations restrictions, so exacerbating the risk of incurring overfitting issues~\cite{marcus2018}.
The proposed approach improve the effectiveness of a sequence-tagging model in these scenarios as confirmed by the experimental results in Section~\ref{sec:experiments}.

Furthermore, in our opinion, our proposal enables an effective human-in-the-loop cooperation scheme for the analysis of low-level log data,
where these two different kinds of automated \emph{Artificial Intelligence} (\emph{AI}) tools complement one the other, in the spirit of neural-symbolic AI~\cite{nesy2022}.
Specifically: 
\begin{itemize}
\item on one side, the contextual event-activity mapping learned by the ML-based model for an event $e_{curr}$ is exploited (as a form of \emph{learning for reasoning}) to ease and speed up the exploration of the valid interpretations of $e_{curr}$, by allowing the selection and ranking of these interpretations based on their probability scores; and 
\item on the other side, the feedback returned by the reasoning framework on the actual validity of these interpretations are exploited to adjust the predictions returned by the tagger.\footnote{In principle, one may try to improve the tagger itself by leveraging the feedback provided by the reasoner on the compliance of the tagger predictions with background knowledge (either in the initial training of the tagger, or in subsequent fine-tuning steps, performed over time according to a continual learning strategy). However, for the sake of flexibility and efficiency, we prefer not to use the given behavioral constraints to guide the training of the tagging model, since some of these constraints might only encode contingent operating conditions or even current user's preferences, so requiring the model to be re-trained again every time these constraints are changed.
}
In particular, since, as discussed above, there is no guarantee that an ML model induced from labelled log data learns the conditional probability distribution 
$p(activity |$ $event,$ $trace$-$wide$ $context)$ accurately (taking the structure of valid interpretations into full account), it is reasonable to expect that the predictions returned by this model can be improved by the reasoner.
\end{itemize}

\paragraph{Novelty and significance of the proposal} 
As to the novelty and significance of our current contribution, it is worth remarking that the problem of supporting an interactive exploratory analysis of multiple candidate event interpretations has not been addressed in the fields of process log analysis and process mining. The only exception is the AAF-based reasoning approach proposed in~\cite{paper}, where, however, the opportunity of leveraging sub-symbolic machine learning methods was not considered at all. Such a kind of hybrid, neuro-symbolic, strategy is instead pursued in our proposal, in order to improve the effectiveness and efficiency of event interpretation analyses. 
In fact, to the best of our knowledge, the current work has been the first attempt to leverage a neuro-symbolic approach in a process mining setting.

Moreover, several key points put this proposal apart from existing neuro-symbolic solutions (e.g., \cite{DBLP:conf/nesy/AhmedTCBV23}) combining logical constraints and ML models: 
\begin{enumerate}
    \item The behavioral constraints considered here are allowed to span over both the input and output variable, rather than over the output only. Indeed, in most cases, neuro-symbolic approaches constrain the output of a prediction model by enforcing conditions that the model's structured output must satisfy.
    \item The constraints are enforced at run time, when using the ML model (i.e. the tagger) to suggest likely event interpretations, rather than when training it (as, e,g,, in~\cite{DBLP:conf/nesy/AhmedTCBV23}), to both ensure that the analysis results comply with the constraints and allow the users to use different constraints in different analysis sessions (while reusing the model as is).
    \item The proposed framework provides the user with a
    versatile and expressive process-oriented data querying capabilities that allows her/him to efficiently derive added-value information from raw low-level log data (and background process knowledge).
\end{enumerate}

This original combination of features %(in terms of explanaibility, flexibility, efficiency) 
makes the proposed neuro-symbolic framework a cutting-edge tool for conducting human-in-the-loop, data-driven, process monitoring and analysis in complex real-life engineering applications.
Some more comments on the novelty and significance of this work can be found in Sections~\ref{sec:novelty} and~\ref{sec:conclusion}.

\subsection{Organization}

The remainder of this paper is organized as follows: Section \ref{sec:related} provides an overview of related work on log abstraction and approaches to integrating symbolic reasoning with deep learning, Section \ref{sec:background} outlines the background of our proposal followed by its description in Section \ref{sec:approach} and in Section \ref{sec:experiments} is presented the experimental setup and related results. Finally, Section \ref{sec:conclusion} discusses the implications of our findings and future directions for research.

\section{Related work} \label{sec:related}
From a technical point of view, the interpretation of low-level event instances in terms of the execution of high-level activities is related to \emph{trace/log abstraction} problem, which has been receiving considerable attention in the field of Process Mining.
Comprehensive surveys on this topic are provided in ~\cite{VanZelst2020} and~\cite{DBLP:conf/bpm/Lim023}. 

Essentially, the interpretation and analysis of log data performed in process mining applications involves mapping events to activities. According to~\cite{DBLP:conf/bpm/Lim023}, such a mapping problem has been addressed in the literature at two different levels of complexity and precision. 
The first kind of mapping simply consists in associating events to  activity classes/types (EI-AC) \cite{abs-ActivityPattern-conf,Baier-IS14,10.1007/978-3-319-97490-3_58}, through some suitably-defined mapping rule $\rho: ei \rightarrow ac$, where each event instance $ei$, represented by some event attributes, is made correspond to an activity class $ac$. 
A more precise mapping is obtained by associating event instances to specific activity instances (EI-AI) \cite{abs-ActivityPattern-conf,conf/emisa/MannhardtT17,abs-ActivityPattern,Ferreira14,deleoni*20,rcis2016-Maikel-sensor,GRA09}, which is a fundamental step in standard process mining pipelines, where log traces are turned into sequences of activity execution steps. This entails defining a mapping rule $\psi: ei \rightarrow ai$, linking each event instance $ei$ to an activity instance $ai$. Here, event instances sharing the same activity class may belong to a single or multiple instances of an activity due to repeated executions.

Naive approaches have been often used to solve the EI-AI mapping by either grouping consecutive event instances with the same activity class as a single activity instance \cite{rcis2016-Maikel-sensor,GRA09} or treating each event instance as a distinct activity instance \cite{BIS15,caise18,abs-ROAD}. 
Alternatively, heuristic approaches allow users to specify conditions for assigning event instances to activity instances, using parameters like the maximum number of event instances per activity instance \cite{Baier-IS14,Baier2018_bis}, the maximum duration of an activity instance \cite{deleoni*20,rcis2016-Maikel-sensor,Baier-IS14,Baier2018_bis}, the number of interleaved activity instances allowed \cite{leonardi2017}, and the maximum duration of these interleaved instances \cite{leonardi2017}.

The rest of this section is structured as follows. 
The first two subsections are devoted to presenting existing knowledge-driven and example-driven approaches, respectively, to supervised log abstraction. 
After briefly illustrating some related neuro-symbolic AI work in the third subsection, we provide the reader with a summary of the main points of novelty of our current proposal with respect to the literature.

\subsection{Knowledge-driven approaches to supervised log abstraction}
\label{sec:knowledge-driven}

An interactive methodology for log abstraction, aiming to transform low-level log traces into activity-level traces, is presented in~\cite{Baier2018,Baier-IS14}. Initially, potential \textit{event-to-activity} mappings are identified by comparing descriptions of activities and event types. Declarative constraints~\cite{declare}, defined at both the event and activity levels using log and process models, are then employed to eliminate unsuitable mappings. However, resolving any remaining ambiguity is left to the user. To distinguish between activity instances within a trace, event clustering is performed, considering instance-border conditions specified by the expert. Notably, both approaches in~\cite{Baier2018,Baier-IS14} rely on the user's expertise to resolve all uncertainties in log trace event interpretation.

The work presented in~\cite{abs-ActivityPattern} takes two types of process models as input: (i) Petri-net-based ``activity patterns", which model the behavior of individual activities in terms of log event types, and (ii) an ``abstraction" model, representing possible execution flows across activities using process-algebra operators. Activity-level abstractions of log traces are obtained by greedily computing the optimal alignments between the traces and a combined DPN model integrating the abstraction and activity pattern models~\cite{alignmentBalanced}.

This approach relies on two key assumptions: (a) the expert can accurately model process behavior using procedural models, and (b) aligning a trace to the expert's specification is sufficient to infer the actual process steps that generated the trace.

The concept of computing multiple event/trace interpretations is also shared with the probabilistic approach in~\cite{nostro-IS18}. This approach utilizes \textit{event-to-activity} mapping and \textit{inter-activity} constraints to eliminate meaningless interpretations. For each trace, a graph-like index structure is employed in~\cite{nostro-IS18} to store all valid interpretations along with their probability scores. However,~\cite{nostro-IS18} does not provide a mechanism for the analyst to efficiently and effectively explore these interpretations.

As in our current work, the interactive analysis of low-level traces, via interpretation queries and explanations, was enabled in \cite{paper} by proposing an ad hoc reasoning framework, based on Abstract Argumentation methods and informed with event-activity mappings and activity constraints. However, this pure-reasoning approach risks falling short when, as observed in \cite{paper} itself, the known behavioral constraints are rather loose, so that interpretation queries are costly to compute and their results are too cumbersome to present. 

In contrast to the previously discussed methods, there are some others in this category rely on simplifying assumptions regarding the mapping between event types and activity types: (i) a \textit{one-to-one} mapping, where each activity type generates events of only one specific type during execution~\cite{Baier-conf14,Baier-conf15}; or (ii) a \textit{many-to-one} mapping, where instances of a given event type can only be generated by one activity type, thus preventing event types from representing ``shared" functionalities across different activities~\cite{Ferreira14}. Additionally, some proposals~\cite{Ferreira14,abs-ROAD,leonardi2017} have limited applicability due to assumptions of no concurrent activity execution or that each activity instance explains only a contiguous subsequence of events within a log trace.

\subsection{Example-driven approaches to supervised log abstraction}
\label{sec:example-driven}

Supervised log abstraction methods~\cite{abs-CRF,tax2018human} assume the availability of labeled log traces, where each step is annotated with the corresponding process activity (and potentially the lifecycle stage). These annotated traces are used to build an abstraction/interpretation model through inductive learning techniques. The underlying assumption is that these examples are sufficient to fully learn the mapping between activities and log events, without the need for explicit symbolic knowledge from the expert.

Specifically, the work in~\cite{abs-CRF} employs linear-chain \emph{Conditional Random Fields} (CRFs) as the model to discover, with the goal of mapping each step in a new trace to one of the reference activities. Similarly, Tax et al.~\cite{tax2018human} use a recurrent neural network (specifically an LSTM) to perform the same event tagging task.

It is worth noting that neither approach in~\cite{tax2018human} nor~\cite{abs-CRF} leverages domain knowledge from analysts about process/activity behavior. Activity-related constraints cannot be easily incorporated into the LSTM-based method in~\cite{tax2018human} or the CRF-based one in~\cite{abs-CRF}. These constraints are difficult to enforce when labeling individual events because neither model considers the activity labels of other events within the same trace.

\subsection{Approaches to integrating symbolic reasoning with deep learning}

In general, neuro-symbolic AI combines sub-symbolic learning methods with symbolic reasoning methods (usually more data efficient and interpretable than the former, but more computationally expensive when it comes to deal with large search spaces and/or uncertainty in the data). This combination can potentially reduce computation costs and times, and make analysis results explainable and traceable.
In a sense, neuro-symbolic systems combine two kinds of reasoning occurring in human brain: type-1 reasoning (fast and intuitive like the pattern recognition tasks accomplished by ML models) and type-2 reasoning (similar to the slower formal way logic-based systems process symbolic information).

In this context, there has been growing interest of late in incorporating logical constraints into deep learning models to improve their performance, data efficiency, and compliance with background knowledge \cite{DBLP:conf/ijcai/GiunchigliaSL22}. 
This has been studied mainly for structured output prediction (SOP) tasks, where the relationships between output labels can be complex \cite{DBLP:conf/nesy/AhmedTCBV23}.
One possible approach to incorporating logical constraints is through the use of semantic loss functions, which penalize models for producing outputs that violate the constraints~\cite{xu2018}. This approach has been shown to be effective in improving the performance of neural models on various tasks, including natural language processing and computer vision \cite{DBLP:conf/ijcai/GiunchigliaSL22}, especially in setting where few training examples are available.

Another approach was proposed by Ahmed et al. \cite{DBLP:conf/nesy/AhmedTCBV23} who introduced the Semantic Probabilistic Layers (SPLs) to address the challenge of injecting logical constraints into deep neural networks while retaining modularity and differentiability. SPLs combine probabilistic inference with logical reasoning in a clean and modular way, by learning complex distributions and restricting their support to solutions satisfying the constraints. In the context of structured output prediction (SOP), SPLs can be used as a drop-in replacement for common predictive layers, guaranteeing that the predictions comply with pre-defined symbolic constraints. 
In~\cite{DBLP:conf/nesy/AhmedTCBV23} this is achieved by leveraging probabilistic circuits, which are computational graphs that can represent both functions and distributions.

\subsection{Comparison with related work}\label{sec:novelty}
If excluding \cite{paper}, there has been no previous attempt to face the specific problem of supporting the interactive (online) exploratory analysis of multiple candidate event interpretations. 
However, the pure (AAF-based) reasoning framework introduced in \cite{paper} to solve this problem, completely disregarded the opportunity to combine this symbolic reasoning module with a data-driven one discovered with the help of machine learning methods --in order to make this kind of interactive analysis more effective and efficient, as proposed in our current work. 

More generally, to the best of our knowledge, this work has been the first one to adopt a neuro-symbolic approach to the activity-level analysis of low-level process traces.
Technically, in our proposal this is done by exploiting symbolic behavioral constraints, embedding the expert's process knowledge, to restrict the support of the learned event-activity distribution to the cases that satisfy the constraints. 
Though there exist neuro-symbolic approaches, like \cite{DBLP:conf/nesy/AhmedTCBV23,DBLP:conf/ijcai/GiunchigliaSL22,xu2018}, combining logical constraints and ML models, our proposal exhibits a number of original aspects that make it different from them: 
\begin{enumerate}
    \item Differently from structured-prediction scenarios like those considered in ~\cite{DBLP:conf/nesy/AhmedTCBV23,DBLP:conf/ijcai/GiunchigliaSL22}, in our problem setting one needs to deal with constraints spanning over both the input (i.e. the past events of the traces under analysis) and output (i.e. the current event of the trace) --rather than over a (multi-dimensional) output only. 
    \item Our behavioral constraints are enforced when using the data-driven model (i.e. the trace tagger) at run time, in an online interpretation session, rather than when training it. This choice is meant to pursue two complementary goals: \emph{(i)} providing the user with high-quality trustworthy information (interpretation-query results and explanations are guaranteed to comply with the constraints), and \emph{(ii)} ensuring a good flexibility-vs-efficiency trade-off (the constraints can vary in different analysis sessions, without requiring the model to be retrained whenever this happens\footnote{This allows for reusing the model as is, under the reasonable assumption that the it captures enough information on the general behaviors of the process under analysis, and that the  constraints being modified characterize specific variants, or use cases, of the process.}).
    \item The proposed framework allows the user to derive added-value information efficiently, starting from raw low-level traces and background knowledge, leveraging an ad hoc, rich, versatile and uncertainty-aware (in the sense of defeasible reasoning) query language. This is not the case of typical neuro-symbolic approaches to informed/constraint-aware machine learning, which usually address simpler inference/prediction problems ---comparable to those of estimating the probability of a single-event interpretation or generating a sample of possible trace interpretations, in our reference setting.
\end{enumerate}

The combination of nice original features (in terms of explanaibility, flexibility, efficiency) summarized above makes our proposal neatly different from existing solutions. 
In our opinion, the proposed neuro-symbolic framework for the human-in-the-loop online log analysis  advances the state of the art in the field of process mining, and can unleash the full potential of log-driven process monitoring and analysis in complex real-life engineering applications.

\section{Background}\label{sec:background}

\subsection{Domain knowledge for model-driven interpretation of low-level event data}
\label{sec:knowledge}
Two kinds of domain knowledge were proposed to be used in \cite{paper} in order to restrict the space of valid interpretations of log events in terms of a given set $\A$ of high-level activity types: 
a set of (many-to-many) candidate type-level mappings between the activities of $\A$ and the event types $\E$ occurring in the process log;
a declarative process model consisting of a number of execution constraints defined over the activities. 

\bigskip
\begin{definition}[Type-level mapping and associated functions]\label{def:mapping}
A \emph{type-level mapping} $\M$ is a relation of the form $\M=\{ (e,a,t) \mid e \in \E, a \in \A, t \in S \}$, where $\E$ is the set of event types occuring in the process log, $\A$ is the set of activity types, and $S=\{  \emph{first}, \emph{intermediate},$ $\emph{last},  \emph{first\&last} \}$ is the set of life-cycle step types characterizing the execution stage of an activity instance. 
Each triple $(e,a,t)$ in $\M$ is meant to encode background knowledge on the fact that an event of type $e$ can be generated as a step of type $t$ of an instance of activity $a$. 
Based on this mapping, one can define the following functions (as done in \cite{paper}): 
$\emph{cand-act}:\E \rightarrow 2^{{\small\A}}$
mapping each event type $e \in \E$ to the set of activities whose execution is known to possibly generate an instance of the same event type as $e$ as one of its steps; 
$\emph{cand-steps}(e) :\E \rightarrow 2^{{\small\A}\times S }$
mapping each event type $e$ to a set of pairs of the form $(a,s) \in \A \times S$ stating that an event of type $e$ can be a step of type $s$ of an execution of activity $a$. 
\hfill $\Box$
\end{definition}

\bigskip
\begin{definition}[Declarative process model]\label{def:process}
A \emph{declarative process model} $W$ over the set $\A$ of activity types is a triple of the form $W=\langle StartAct, maxInst, IC \rangle$, where: 
\begin{itemize} 
\item $StartAct \subseteq \A$ is a set of starting activities (i.e., the activities that are allowed to be executed at the beginning of the process)
\item $maxInst$ is a function associating each activity $a \in \A$ with a positive integer (or  $\infty$, in case the information is not available) indicating the maximum number of times that a new instance of $a$ can be started within a process instance; and 
\item $IC$ is a set of temporal rules of the following forms: $MC: A\Rightarrow_T B_1| \dots| B_k$ (\emph{must-constraint}), $NC: A\Rightarrow_T \neg B$ (\emph{not-constraint}),  %or of their respective time-symmetric versions (i.e., positive and negative ``precedence'' constraints). 
$PC: A | \ldots | A_k \Leftarrow_T B$ (\emph{precedence-constraint}), and  $NPC: \neg~A\Leftarrow_T~B$ (\emph{negative precedence-constraint}), 
for some given activities $A,A_1, \ldots, A_k, B, B_1,\dots,B_k$ and time period $T \in [1..\infty]$.
Constraint $MC$ (resp., $NC$) means that an instance $A$ must (resp., cannot) be followed by an instance of one of the activities $B_1, \dots, B_k$ (resp., of activity $B$) in $T$ steps; analogously, constraint PC  (resp., $NPC$) means that every instance of $B$ must  resp., must not) be preceded by an activity instance of a type $Y  \in \{A_1,\ldots,A_k\}$ (resp., of type $A$) such that the number of steps between the last event of $Y$ and the first event of $B$ is at most $T-1$.
\hfill $\Box$
\end{itemize}
\end{definition}

\medskip
\begin{example}[contd.]\label{ex:example2}
Consider the toy scenario of Example \ref{ex:example}, and assume that candidate mapping between event types and activities presented there can be refined through the following life-cycle relationships (represented according to the notation of Def.~\ref{def:mapping}):  
$(e_1, B, first)$, $(e_2, B, intermediate)$, 
$(e_2, B, last)$, $(e_3, B, last)$, 
$(e_3, A, first\&last)$, $(e_3, A, first)$, $(e_4, A, intermediate)$, $(e_4, A, last)$, $(e_5, A, last)$, $(e_5, C, first)$, 
$(e_6, C, last)$ and $(e_6, C, first\&last)$.

Let us also assume that all possible executions of the process are ensured to comply with the following constraints:
(i) both activities $B$ (preparation) and $C$ (pre-surgery) are executed at most once in every process instance, while activity $A$ (hospitalization) could be executed twice in a process instance;  
(ii) every execution of activity $A$ must be 
followed immediately by an execution of activity $C$ (pre-surgery), \pezzonuovo{and 
every execution of activity $B$ must be followed in at most three steps by an execution of activity $C$}; and 
(iii) every trace starts with (an instance of) activity $B$.
In our framework, this can be modeled with a declarative process model $W=\langle StartAct, maxInst, IC \rangle$ such that: $StartAct = \{B\}$, $maxInst(B)=maxInst(C)=1$, $maxInst(A)=2$, and $IC =\{MC_1: A \Rightarrow_1 C, MC_2: B \Rightarrow_3 C  \}$.

Clearly, of the two candidate interpretations of trace 
$\trace=[e_1,e_2,e_3,e_4,e_5,e_6]$ of Example \ref{ex:example}, only interpretation $I_2$ (i.e. ``\emph{$e_1,e_2$ is the sequence of steps produced by an instance of $B$, while
$e_3,e_4$ and $e_5,e_6$ are the step sequences produced by instances of $A$ and of $C$, respectively}'') 
is admissible, while the other interpretation $I_1$ (i.e. ``\emph{$e_1,e_2$ is the sequence of steps produced by an instance of $B$,
$e_3,e_4,e_5$ is the sequence of steps produced by an instance of $A$ and $e_6$ is the unique step of an instance of $C$}'') 
can be disregarded \pezzonuovo{as it violates the constraint that
every execution of activity $B$ must be followed in at most three steps by an execution of the activity $C$. More in detail, in $I_1$ the first step of $C$ is step $6$, while the last step of $B$ is step 2, which means that $A$ occurs $4$ steps after $B$}.
\hfill $\Box$
\end{example}
\medskip

\subsection{An approach to low-level event interpretation based on Abstract Argumentation}
\label{sec:aaf}

\paragraph{Abstract Argumentation Frameworks (AAFs)}
An AAF can be viewed as a graph where nodes and edges model  \textit{arguments} and \textit{attacks} from an argument to another, respectively.
Given argument set $S$ and argument $\alpha$, we say that ``\emph{$S$ attacks $\alpha$}'' (resp., $\alpha$ attacks $S$)
if there is an argument $\beta$ in $S$ such that $\beta$ attacks $\alpha$ (resp., $\alpha$ attacks $\beta$).
Given arguments $\alpha$, $\beta$ such that $\alpha$ attacks $\beta$, 
an argument $\gamma$ is said to ``\emph{defend $\beta$}''  if $\gamma$ attacks $\alpha$.
Argument set $S$ is said to ``\emph{defend $\beta$}''  if $S$ contains some argument defending $\beta$.
Argument \emph{$\alpha$ is acceptable w.r.t. $S$} if every argument attacking 
$\alpha$ is attacked by $S$, while $S$ is \emph{conflict-free} if there is no attack between its arguments.
Different semantics have been proposed 
to identify ``reasonable'' sets of arguments, called \textit{extensions}~\cite{Dung95}, in an AAF.
In particular, a set $S\subseteq A$ is said to be 
an \textit{admissible extension} 
iff $S$
is conflict-free and all its arguments are acceptable w.r.t. $S$.
An admissible extension that is maximal (w.r.t. $\subseteq$) is a
\textit{preferred extension}. 
Acceptance of an argument $\alpha$ in an AAF $F$ can be stated under both skeptical and credulous perspectives. In particular, adopting the preferred semantics (as in \cite{paper}), 
$\alpha$ is \textit{credulously accepted} (resp., \textit{skeptically accepted}), 
if it belongs to at least one (resp., every) preferred extension of $F$.

\paragraph{The core interpretation problem and AAF-based solution approach to it}
Given a running trace $\trace$, as soon as a new event $\ecurr$ appears in $\trace$, in \cite{paper}, 
the analyst is allowed to ask for $(a)$ answers to \textit{Interpretation Queries} on the alternative \emph{valid} interpretations of $\ecurr$, 
and $(b)$  \emph{explanations} for certain interpretations considered \emph{not valid}, where interpretation validity descends from 
well-structuredness properties and given domain knowledge (described later on).
Basically,  interpretation queries allow for asking whether $\ecurr$ can be viewed as a step of an instance of a given activity $A$, 
possibly stating conditions on the activity lifecycle  (e.g.,``\emph{is $\ecurr$ the \underline{initial} step of an instance of $A$}?" 
or on the activity occurrence (e.g., ``\emph{is $\ecurr$ a step of the \underline{$2$nd} instance of $A$?}"). 
Any such a query can be evaluated under \emph{skeptical semantics}, in order to know whether the proposed interpretation 
is the \emph{unique} valid one for $\ecurr$.
Moreover, \emph{enumeration queries} ask for all valid interpretations of $\ecurr$ satisfying the conditions required.

Technically, the core problem of computing valid (w.r.t. domain knowledge) interpretations of events in a trace $\trace$ is modelled as a dispute 
via an ad hoc AAF $F(\trace)=\langle$ \emph{Arg}, \emph{Att} $\rangle$, 
on top of which interpretation queries and explanation requests can be answered.
$F(\trace)$ is built incrementally by adding arguments of the following kinds (and attacks between them) as a new step $\trace[curr]$ of $\trace$ arrives (where $curr \in \mathbb{N}$ denotes the index of the last step of $\trace$): 

\medskip
\noindent
1) \underline{\emph{Interpretation arguments}} 
of the form $\langle e_i, A, X, j \rangle$,   
where $A$ is an activity label, $j$ is a positive integer and $X\in\{ \ \emph{first}, \ \emph{intermediate}, \ \emph{last},  \ \emph{first\&last} \ \}$ 
indicates a life-cycle execution stage for $A$.
Such an argument encodes the interpretation of $e_i$ as a step of type 
$X$ of the $j$-th execution instance of activity $A$. 
Notably, the incremental generation of candidate interpretation arguments for any newly arrived step $\trace[curr]$ simply relies on using the functions \emph{cand-act} and \emph{cand-steps} defined in Def.~ \ref{def:mapping}, which look at event type of $\trace[curr]$ and at the underlying set $\M$ of known type-level mappings between events and activities.

\medskip
\noindent
2) \underline{\emph{Undermining arguments}} of the following different forms:
\begin{itemize}
%$(i)$ 
\item \emph{NotInterpreted}$_i$, meaning that no interpretation has been chosen for step $\trace[i]$.
%$(ii)$ 
\item \emph{NotEnoughExecutions}$_\alpha$, where $\alpha=\langle e_i, A, X, j \rangle$ with  $X\in \{$\emph{first}, \emph{first\&last}$\}$, meaning that $\alpha$'s interpretation of $e_i$ as the start of the $j$-th instance of $A$ 
contrasts with interpreting no previous step as the start of the $(j\!-\!1)$-th instance of $A$.
%$(iii)$ 
\item \emph{NotStarted}$_\alpha$, where $\alpha=\langle e_i, A, X, j \rangle$ with  $X\in \{$\emph{intermediate}, \emph{last}$\}$, 
meaning that $\alpha$'s interpretation of $e_i$ as a non-initial step of the $j$-th instance of $A$ 
contrasts with interpreting no previous step as the start of the $j$-th instance of $A$.
%$(iv)$ 
\item \emph{NotEnded}$(j,A)$, where $j$ is an instance counter and $A$ an activity, meaning that the trace has terminated and some event was interpreted as the initial step of the $j$-th instance of $A$, but no event has been interpreted as the last step of this instance.
%$(v)$ 
\item $MC_q^i$, descending from some must-constraint $MC_q: A \Rightarrow_T B_1| \dots| B_k$, and meaning that the $i$-th event of $\trace$ cannot be 
interpreted as the final step of an instance of $A$, for no subsequent event (in $T$ steps) has been interpreted as the start of an instance of some $B_i$ with $i \in \{1,\ldots,k\}$.
%$(vi)$ 
\item $PC_q^i$, for each positive precedence constraint $PC_s: A_1 | \ldots | A_k \Leftarrow_T B$, meaning that the $i$-th event of $\trace$ cannot be 
interpreted as the initial step of an instance of $B$, for no previous event (in $T$ steps) has been interpreted as the end of an instance of some $A_i$ with $i \in \{1,\ldots,k\}$. 
%$(vii)$ 
\item Further attacks are included to guarantee that $\emptyset$ is the only admissible extension of $F(\trace)$ iff there is no valid interpretation of $\trace$ (i.e. trace behavior deviates from the process model). This is the case of self-attacks on all \emph{NotInterpreted}$_i$ 
\emph{NotEnoughExecutions}, \emph{NotStarted}, \emph{NotEnded}.
and the special attack from \emph{NotInterpreted}$_1$ towards each argument of type 
\end{itemize}

\begin{figure}[!t]
	\includegraphics[width=0.85\textwidth]{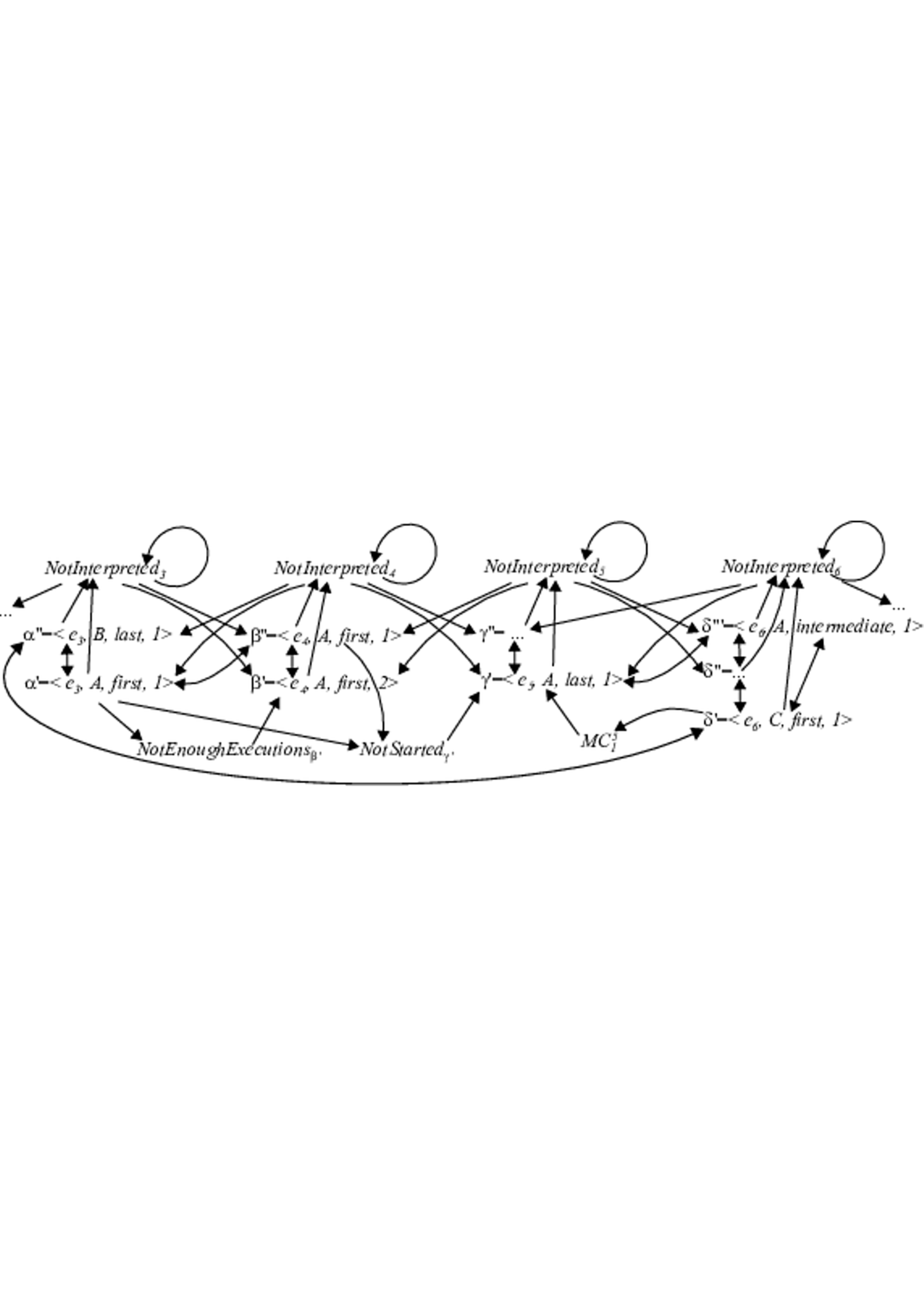}
	\centering
	\caption{An excerpt of the AAF $F(\trace)$  (edges with two arrows represent pairs of mutual attacks) for the toy interpretation scenario described in Examples~\ref{ex:example},~\ref{ex:example2} and ~\ref{ex:example3}. Many of the arguments of the AAF (including, e.g., the interpretation arguments $\tuple{e_4, A, last, 1}$) are not shown here, for the sake of readability.  }
	\label{fig:framework}
\end{figure}

\medskip
\begin{example}[contd.]\label{ex:example3}
Consider again the toy event interpretation scenario described in Examples~\ref{ex:example} and~\ref{ex:example2}, with the partial trace $\trace = [e_1, \ldots, e_6]$ and its associated interpretations presented there. 
It is easy to see that interpretation $I_2$ can be encoded in an AAF framework of the form defined above through the interpretation arguments {%\small
$\tuple{e_1, B, first, 1}$, $\tuple{e_2, B, last, 1}$,$ \tuple{e_3, A, first, 1}$, $\tuple{e_4, A, last, 1}$, $\tuple{e_5, C,first,1}$, and $\tuple{e_6, C,last,1}$}.

An excerpt of the AAF resulting from processing the above-mentioned trace $\trace$ is depicted in Fig.~\ref{fig:framework} --many of the arguments of the AAF (including some of the interpretation arguments mentioned above, like $\tuple{e_4, A, last, 1}$) are not shown here, for the sake of readability.
The AAF  includes several conflicting event interpretations, encoded by interpretation arguments with mutual attacks, e.g.: 
$\alpha'$ and $\alpha''$ (representing alternative interpretations of the same event),
$\alpha'$ and $\beta''$ (interpreting distinct events as first steps of the first instance of $A$), and 
$\gamma'$ and $\delta'''$ (where $\delta'''$ interprets $e_6$ as a step of an instance interpreted as closed by $\gamma$).
As an example of how the AAF ensures activity counters to be consistent, consider \emph{NotEnoughExecutions}$_{\beta'}$, which attacks $\beta'$ 
and it is counter-attacked by $\alpha'$. 

Importantly, the AAF enforces the choice of an interpretation for each trace step. In particular, it contains exactly one argument \mbox{\emph{NotInterpreted}$_x$} for each step $e_x$, 
that is attacked by the interpretation arguments over the same step $e_x$ and attacks the 
arguments over the immediately preceding and following steps $e_{x-1}$ and $e_{x+1}$.

The AAF also ensures that activity instances can be prolonged only if previously started.
For example, in Fig.~\ref{fig:framework}, $\gamma'$ interprets $e_5$ as a non initial step of the first instance of $A$ and is attacked by \emph{NotStarted}$_\gamma'$, and is 
defended from this attack by $\alpha'$ and $\beta''$, that interpret some previous steps as the starting steps of the same instance of $A$.

All the constraints in the given process model are 
enforced as well.
For example, in Fig.~\ref{fig:framework}, the must-constraint
$MC_1: A\Rightarrow_1 C$ (triggered by argument $\gamma'$ over step $5$)
is encoded by the undermining argument $MC_1^5$ and the attacks $(MC_1^5, \gamma')$
and $(\delta',MC_1^5)$.
\hfill $\Box$
\end{example}
\medskip

It was proven in \cite{paper} that there is a biunivocal correspondence between
the set of valid interpretations of $\trace$ and the set of preferred extensions of $F(\trace)$, 
so one can reason on the interpretations of the current event $\ecurr$ by deciding the 
acceptance of interpretation arguments over $\ecurr$.
We refer the interested reader to \cite{paper} for details on the computation of the AAF $F(\trace)$ and of answers to interpretation queries and associated explanation requests.

\section{Proposed approach: combining AAF-based and ML-based interpretation}\label{sec:approach}
Let $\A$ denote the set of the reference process activities, and $\E$ be the set of event types that may be generated by the process under analysis.
Let $\U$ denote the universe of all possible log events, and $e$-$type: \U \rightarrow \E$ be a function mapping every event $e \in \U$ to the respective event type $e$-$type(e) \in \E$.

Our approach to the explorative analysis of the valid interpretations of any event $e_{curr}=\trace[curr]$ of a process trace $\trace$ consists in combining two different AI tools: 
$(i)$ an \emph{AAF-based reasoner} $R$ instantiating an extended version of the AAF-based framework defined in \cite{paper}; and 
$(ii)$ a \emph{trace tagger}, which is meant to be trained over a collection of per-step annotated log traces to be used incrementally to map $e_{curr}$ to the activities in $\A$ probabilistically.

Before explaining how we propose to exploit these two components, in a complementary synergistic way in the analysis of low-level log data, let us summarize some major features of these components in the following two definitions, respectively.

\bigskip
\begin{definition}\label{def:reasoner} 
An \emph{AAF-based reasoner} (or simply \emph{reasoner} for short) $R$ is a knowledge-driven model instantiating the AAF-based framework of \cite{paper}. This model is assumed to have been equipped with the two kinds of domain knowledge described in Section \ref{sec:knowledge}, i.e. a set $\M$ of type level mappings between event types and activity types, and a declarative process model $W$ specified in terms of the given activity types.
A reasoner $R$ is also assumed to implement three different functions, which can be used 
interactively as discussed in \cite{paper}:
\begin{itemize}
\item A function $R$.\texttt{answer}($\cdot$) taking a given interpretation query of the form $\alpha = \langle e_{curr}, a, s, j \rangle$ as input and returning an answer to 
the query, where $a \in \A$, $j \in \mathbb{N} \cup \{ \_ \}$, and $s \in S \cup \{\_ \}$, $S = \{$ \emph{first}, \emph{intermediate}, \emph{last},  \emph{first\&last} $\}$, 
and $\_$ is a placeholder meaning that no actual condition is expressed on an interpretation field (i.e., either the life-cycle step or the activity instance number, or both).
In case the given interpretation query $\alpha$ contains no occurrence of placeholder $\_$, the query is considered as a boolean query, and the result $R.\texttt{answer}(\alpha)$ 
returned for it can either be \texttt{YES}, (if $\alpha$ represents a valid interpretation of event $e$) or \texttt{NO} (otherwise); 
in case $\alpha$ contains instead some occurrence of placeholder $\_$, the result $R.\texttt{answer}(\alpha)$ consists of all the valid interpretations of event $e$ that match 
the conditions expressed in $\alpha$ (while possibly taking whatever value on the fields marked with $\_$ in the query).
\item
A function $R$.\texttt{explain}($\cdot$) which returns a (possibly empty) list of invalidity reasons for a given interpretation query of the form $\alpha = \langle e_{curr}, a, s, j \rangle$ provided as input to the function, where $a \in \A$, $j \in \mathbb{N} \cup \{ \_ \}$, and $s \in S = \cup \{\_ \}$.
The list of invalidity reasons is not empty only when $\alpha$ is invalid (i.e., it is reckoned by the reasoner not to model any valid interpretation of $e$), and hence it makes sense for the user to ask the reasoner for explaining this result.
\item 
Two functions $R$.\texttt{getAAF}()  and $R$.\texttt{setAAF}($\cdot$) that allow for accessing and overwriting, respectively the Abstract Argumentation Framework (AAF) $F(\trace)$, stored internally to $R$;
\item
A function $R$.\texttt{updateAAF}($\cdot$) which allows for incrementally extending the AAF $F(\trace)$ with information concerning the newly arrived trace step $e_{curr} = \trace[curr]$. 
Specifically, this function takes as input a set $A$ of candidate activities for $e_{curr}$, and then extends AAF with: (i) a number of interpretation arguments corresponding to different interpretations of $e_{curr}$ as an event generated by (some instance of) an activity in $A$, and (ii) inserting a number of auxiliary arguments and attacks guaranteeing the consistency of the interpretations that will be eventually accepted by the reasoner.
In our approach, we will call $R$.\texttt{updateAAF}($A$) in two different ways:
\begin{itemize}
\item 
by providing it with all the activities linked to the eventy type of $e_{curr}$ in the given set $\M$ of type-level mappings between events and activities, i.e. by instantiating $A=$\emph{cand-act(e-type}$(e_{curr}))$ (cf. Def. \ref{def:mapping}); or
\item
by providing it with a (local, context-dependent) set $A \subseteq \A$ of candidate activities for $e_{curr}$, which is meant to override the set $cand$-$act(e_{curr})$ mentioned above.
\hfill $\Box$
\end{itemize}
\end{itemize}
\end{definition}

\noindent 
The latter use case of function $R$.\texttt{updateAAF} allows the reasoner to generate the candidate interpretations of $e_{curr}$ in a more informative way (by possibly exploiting information concerning both the previous steps of the trace and the different data properties of $e_{curr}$, beside the very event type). 
In order to allow the reasoner (through function $R.update$) to generate an informative (context-dependent) set of candidate interpretation arguments for the current trace step $e_{curr}=\trace[curr]$ --rather than just derive these arguments from given global mappings between event types and activity types, as done in \cite{paper}--- we here propose to resort to a specific kind of sequence tagging model, which is defined below.

\bigskip
\begin{definition}\label{def:tagger} 
A \emph{trace tagging model} (or simply, \emph{tagger}, for short) $M$ is a sub-symbolic ML model \footnote{In the current implementation of our approach and in the experimentation, two alternative neural network architectures have been used to instantiate our trace tagging models} (suitably trained over activity-annotated log traces) that can incrementally predict which activities could have generated the current $e_{curr}$ of any ongoing
trace $\trace$ under analysis.
Such a model $M$ is required to implement 
%the following functions: 
%\begin{itemize}
%\item 
A probabilistic event tagging function $M.\texttt{predict}(\cdot): \U \rightarrow  \Delta_{\A}$ that maps any current step $e_{curr}=\trace[curr]$ to a probability distribution $M.\texttt{predict}(e_{curr}) \in \Delta_{\A}$ over the activities, where $\Delta_{\A} \subseteq [0,1]^{|\A|}$ denotes the space of all probability distributions over set $\A$, based on $e_{curr}$ (and possibly the events preceding it if $M$ is a sequence-oriented model).
Moreover, to allow $M$ to conveniently exploit the different kinds of event data fields stored in the 
log traces, it can be equipped with an event embedding layer.\footnote{
In the current implementation of the approach, this layer can be configured in a way that the categorical fields of an event (including the event type) can be mapped to either one-hot representations or automatically trained field embeddings, while each numerical event field is left as is (modulo min-max normalization).}
The so-obtained representations of the fields of each event $e$ are simply concatenated one the other, to produce a representation for $e$ as a whole (of a form that an be taken as input by typical machine learning models, including neural networks).
\hfill $\Box$
\end{definition}

The rest of this section is devoted to illustrating how the two models $R$ and $M$ described before are used in our framework to support: $(i)$ the interpretation-oriented analysis of new log trace, and $(ii)$ the training of $M$ over (annotated) historical log traces.

\begin{figure}[t]
	\includegraphics[width=\textwidth]{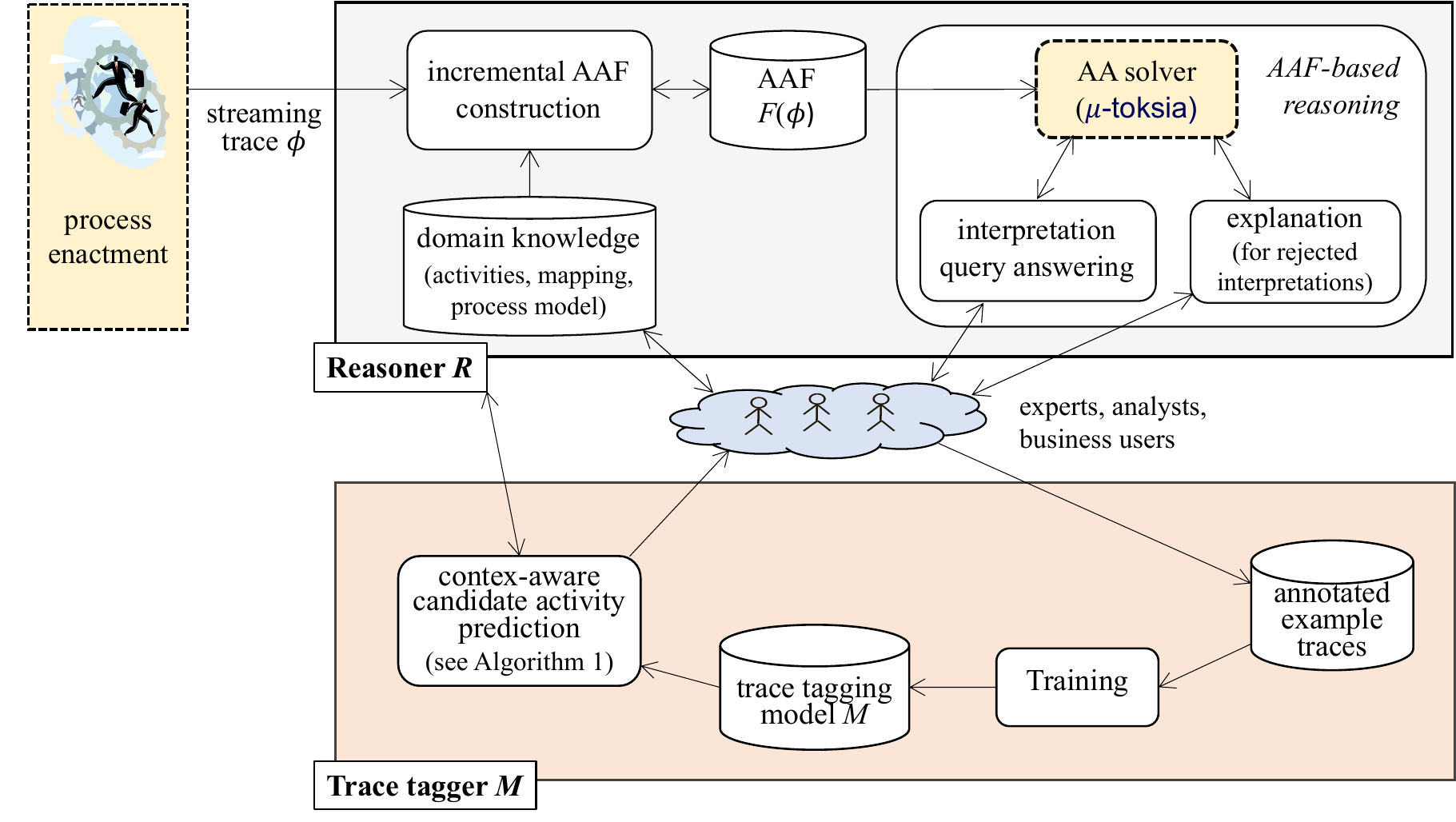}
	\caption{Conceptual architecture of the proposed framework.}
	\label{fig:tool}
\end{figure}

A high-level conceptual view of the different functionalities implemented by the framework is sketched in 
Figure \ref{fig:tool}.

\floatname{algorithm}{Algorithm}
\algsetup{indent=2em}
\begin{algorithm}[htbp]
	\caption{Interactive online analysis of running trace $\trace=[e_1, \ldots, \ecurr]$, after the arrival of any new event $\ecurr$ for the trace}
	\begin{algorithmic}[1]
	\REQUIRE 
	current event $\ecurr$; 
	a trace tagger $M$, trained on annotated traces (see Section \ref{sec:training}); 
 a reasoner $R$ (Def.\ref{def:reasoner}), instantiated with a global type-level mapping $\M \subseteq \E \times \A \times S$ (Def. \ref{def:mapping}) and a declarative process model $W$ (Def. \ref{def:process}), where $\A$ denotes the universe of activities;
	    max number $k \in [1 .. |\A|]$ of candidate activities to consider for \ecurr; 
    Laplace-like smoothing factor $\gamma$;	
    an AAF $F(\trace)$, associated with $R$, updated incrementally at all previous trace steps $\trace[1], \ldots, \trace[curr-1]$;
	
		\textit{// Keep a copy of the AAF $F(\trace)$ and update it as in \cite{paper}}
		\STATE \emph{oldF} = $R$.\texttt{getAAF()}\\%[2pt]
		\STATE $R$.\texttt{updateAAF}\emph{(cand-act(e-type(\ecurr))};\\[4pt]
		
		\textit{// Use the tagger $M$ to predict which activities could have generated \ecurr: the result is a categorical distribution mapping each activity in $\A$ to a probability value $pd[a]$ in $[0,1]$.}
		\STATE $pd:= M.\texttt{predict}(\ecurr)$;\\[4pt]
		
		\textit{// Filter out all invalid predictions (i.e., those conflicting with domain knowledge), and smooth and normalize the remaining probability values.}
		\FOR{$a \in \A$}
		\IF{$R$.\texttt{answer}($\langle \ecurr, a, \_, \_ \rangle$) $= \emptyset$}
		%		\FOR{$\beta\in\altint(\alpha)$ }
		\STATE $pd$[a]:=0;\\[4pt]
		\ELSE
		\STATE $pd$[$a$]:= $\frac{pd[a]\cdot N + \gamma}{N+\gamma \cdot |A|}$; \emph{// Smooth the probabilities of valid candidates}
		\ENDIF
		%		\ENDFOR
		\ENDFOR	
		\STATE 
        normalize the probability scores of the elements in $pd$ in a way that they sum up to 1;
        \\[4pt]
	
		\textit{// Recompute the AAF using only the interpretations of $\ecurr$ that that comply withe the given domain knowledge.}\\
		\STATE $R$.\texttt{setAAF}(\emph{oldF});\\%[2pt]
		\STATE let $A$ be the set of activities (i.e., candidate interpretations of $\ecurr$) associated with non-zero probability in $pd$;
		\STATE $R$.\texttt{updateAAF}($A$);\\[4pt]
		
		\textit{// Support interactive analyses of \ecurr interpretations.}\\
		\STATE present the top-$k$ activities in $pd$ ranked based on probability scores;\\%[2pt]
		\STATE allow the user to pose interpretation queries and explanation requests over $\ecurr$, via functions $R$.\texttt{answer}($\cdot$) and $R$;\texttt{explain}($\cdot$), respectively.\\[4pt]		
	\end{algorithmic}
	\label{algo:analysis} 
\end{algorithm}

%\end{landscape}

\subsection{Online interpretation of low-level log traces}\label{sec:analysis}
The proposed core processing scheme supporting the online analysis of low-level log traces is sketched in Algorithm \ref{sec:analysis}, in the form of pseudo code.
The algorithm is assumed to be invoked in an incremental fashion, over a given log trace $\trace$ under analysis, every time an event, referred to as $e_{curr}=\trace[curr]$, becomes available that corresponds 
to a novel execution step of the trace.
As a prerequisite to such an invocation, we assume that a reasoner $R$ and a trace tagging model $R$ have been instantiated and configured suitably, of the forms described in Definitions \ref{def:reasoner} and \ref{def:tagger}, respectively.
In particular, the reasoner $R$ is assumed to store given global type-level mapping $\M$ (between event types and activity types, as in Definition \ref{def:mapping}) and declarative process model (of the form defined in Definition \ref{def:process}), as well as to maintain an Abstract Argumentation Framework (AAF) $F(\trace)$, which has been updated incrementally so far as specified in the algorithm itself.

The algorithm starts updating the AAF $F(\trace)$ maintained by $R$, to allow it to reason on possible interpretations of the current step $\ecurr$ of the trace $\trace$, after keeping apart a copy of the AAF 
as it was before this update.
At this stage, the AAF is updated by calling the simplest version of function $R.updateAAF$, which 
relies on extending the AAF on the basis of the interpretation arguments of the form $\langle e$-$type (e_{curr}), a, s \rangle$ that comply with the global type-level mapping $\M$.

As explained in the following, this updated version of the AAF will be replaced with a more informative and more compact one, obtained by refining the former version of the AAF (as it was stored in $oldF$) with local context-aware candidate activities obtained with the help of the trace tagger $M$. Specifically, the new AAF is more compact, as it is derived from the previous one by removing interpretation arguments that either do not correspond to valid interpretations of the current event or have a low probability of being the correct interpretation, as explained in what follows.

To this end, the tagger $M$ is first exploited to obtain an estimate $pd$ of the conditional probability distribution $p(activity\mid \ecurr, \trace[1:curr-1])$, such that $pd[a]$ represents the probability that 
activity $a$ generated $\ecurr$ (given the former steps $\trace[1:curr-1] = [\trace[1], \ldots, \trace[curr-1]]$).
Since model $M$ is not ensured to capture these probabilities in a perfect way, and it may well assign non-zero probability to activities appearing in no valid interpretation of event $\ecurr$, we exploit the reasoning abilities of $R$ (and the underlying updated version of $F(\trace)$) to possibly improve $pd$. 
In particular, for each activity $a$ that is deemed as invalid (w.r.t. domain knowledge) by $R$, the estimated probability is zeroed in $pd$. 
In order to prevent the extreme case where $M$ assigned non-zero probabilities only to invalid activities, the probability values in $pd$ are made undergo a Laplace-like smoothing procedure, which allows for possibly 
assigning non-zero probabilities to activities that were given a probability of 0 by $M$, 
provided that these activities can be mapped to $\ecurr$ according to $\M$. 

After performing the above-described knowledge-driven revision of activity probabilities, these are normalized in a way that the sum up to 1, wo that $pd$ can still be interpreted as a categorical probability distribution. 
The selected activities, stored in a set $A$, are used to recompute the AAF $F(\trace)$ (starting from the saved temporary copy $oldF$ of the AAF, reflecting the situation before the arrival of $\ecurr$).

The selected candidate activities (along with their associated probabilities, suitably normalized) are also presented to the user, in a ranked way (the higher the probability, the higher the rank), to allow her/him to focus on the most probable activity-wise interpretations of $\ecurr$ in interactively posing interpretation queries and explanation requests to $R$.

The parameter $k$ is intended to be set by the user when it is expected that multiple possible activities may correspond to a single event, and the user prefers not to be presented with too many alternatives.
In principle, one could think of enforcing this requirement also in the construction of the AAFs, to speed up the computation of interpretation queries and associated explanations.
However, this may lead to information loss, so that the results returned by reasoner $R$ may be incomplete as in \cite{paper}.
Studying the efficiency-vs-completeness trade-off of using such a sort of beam-search decoding scheme is beyond the scope of this work, and it is left to future research.

\subsection{Instantiating and training the trace tagging model}\label{sec:training}
The tagging model $M$ of the kind defined in Definition \ref{def:tagger} can be trained over a given set 
$L=\{(\trace, \Psi \}$ of activity-annotated log traces, where for each pair $(\trace, \Psi)$ appearing in $L$, $\trace \in \U^*$ is a sequence of events, $\Psi \in \A^*$ 
is a sequence of activity labels of the same length as $\trace$, and for each step $\trace[i]$ of $\trace$, 
$\Psi[i]$ indicates the (ground-truth) activity that generated $\trace[i]$.

\pezzonuovo{
In principle, the task of training of $M$ over such a collection $L$ of annotated traces can be accomplished by 
iteratively predicting the activity that generated the $i$-th step/event $\trace[i]$ of each trace in $L$, while using the sequence $\trace[1:i]=[\trace[1], \ldots, \trace[i]]$ as input and $\Psi[i]$ as the associated ground-truth output ---i.e. the activity label associated with the actual activity that generated $\trace[i]$.
}

\pezzonuovo{
Regarding the output of model $M$ (with as many elements as the process activities) as a categorical probability distribution (obtained by applying softmax normalization to a list of logits),} 
one can simply optimize the parameters of $M$ by using common SGD-based schemes with a classical cross-entropy loss. \footnote{\pezzonuovo{
When it comes to interpret an unseen trace $\phi'$, for each step $\phi'[i]$ of the trace, such a tagging model is expected to return, for each activity $a$, an estimate of the probability that $\phi'[i]$ was generated by an instance of $a$.
}
}

\begin{figure}[htbp]
	\centering
	\begin{tabular}{ccc}
		\hspace*{-5mm}\includegraphics[width=0.50\textwidth]{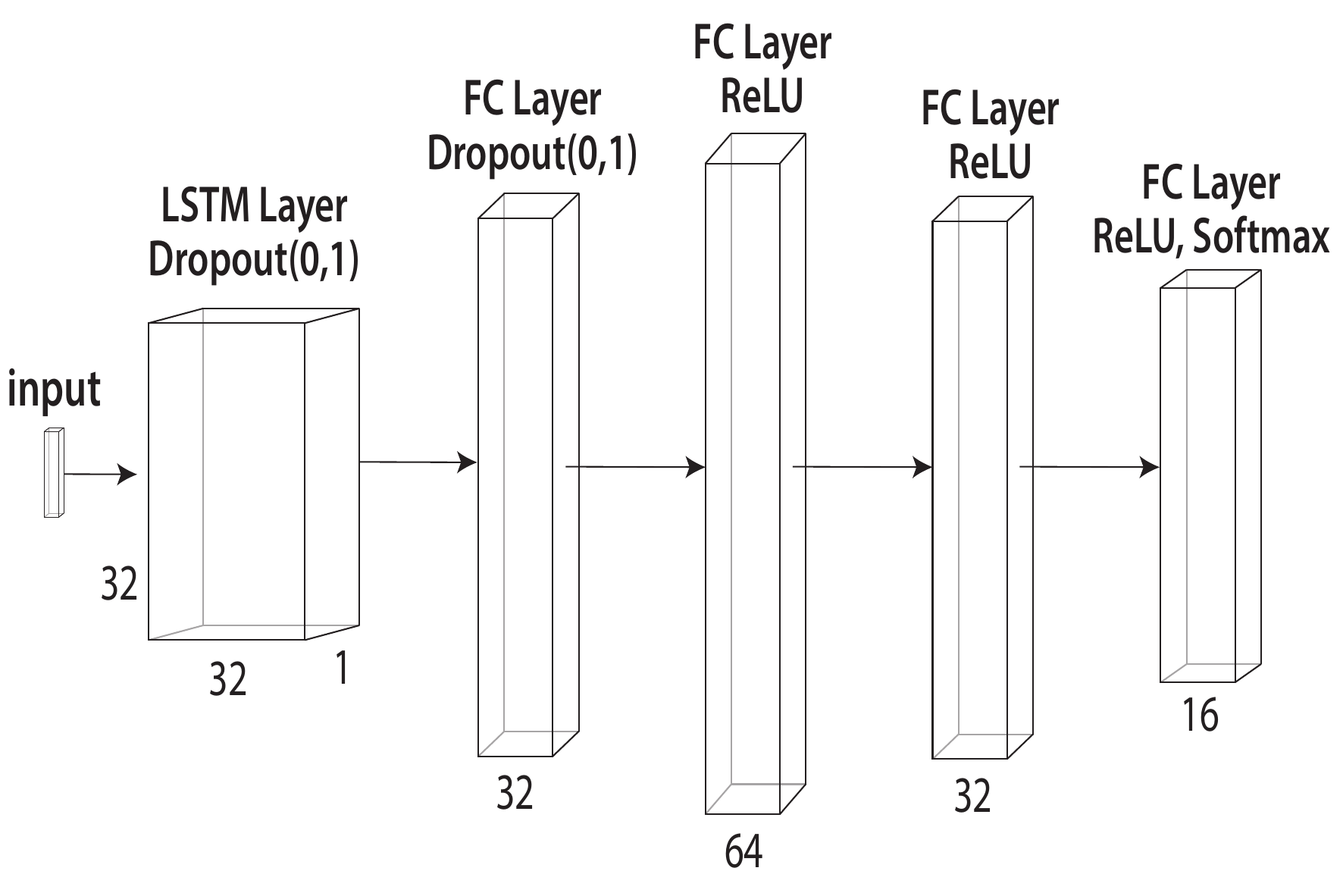} & \ 
		& \includegraphics[width=0.50\textwidth]{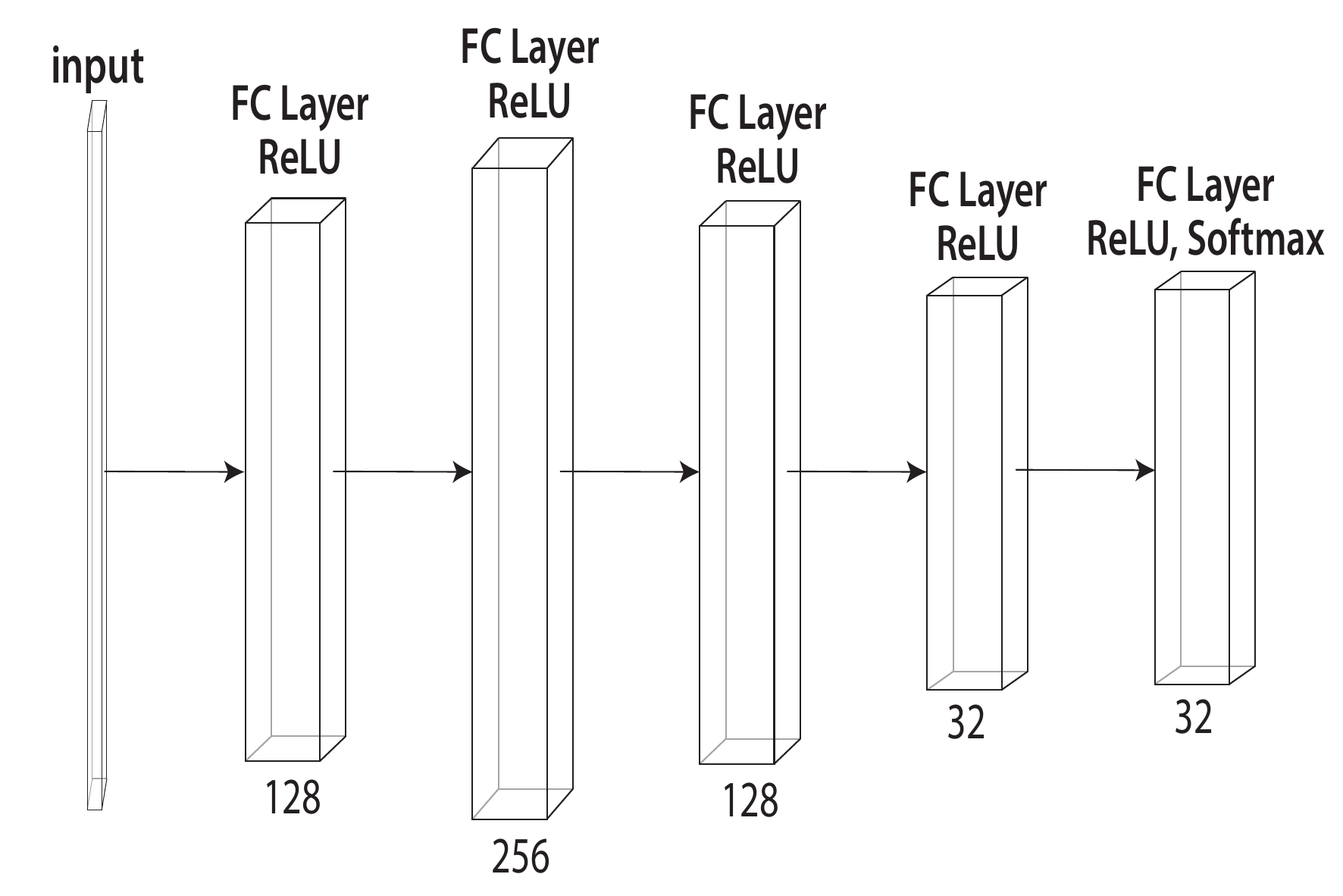} \\
		(a) & & (b)
	\end{tabular}
	\caption{Architectures of \ma (a) and \mb (b).}
	\label{fig:nn}
\end{figure}

To implement the trace tagger model, two simple neural-network architectures, denoted as \ma\ and \mb\, have been considered in the current version of framework proposed. 
These alternative architectures are sketched in Figure~\ref{fig:nn}$(a)$ and $(b)$, respectively.\footnote{In principle, the proposed framework is parametric to the form of the sequence tagging model, which could be implemented by using more expressive architectures (like state-of-the-art ones RNN-based or Transformer-based ones). However, this would increase the risk of incurring severe overfitting issues when training the model on few labeled traces, so providing the user with misleading results.}

As commonly done when making data undergo a neural network, each event is assumed to be converted into a  vector by using some kind of embedding method (in the experiments of Section~\ref{sec:experiments},  word2vec~\cite{word2vec} model was used to this end, after observing that it allowed us to obtain similar results as when using a, more cumbersome, one-hot encoding of the activities).

The structures of the two tagger architectures are described in what follows, in two separate paragraphs.

\paragraph{Tagger model: architecture $\ma$}
\ma\ receives as input an event sequence (where each event is represented by its word2vec embedding)
and processes it through an LSTM layer to capture sequential dependencies. It then passes the LSTM output through fully connected layers, reducing the dimensionality and introducing non-linearity. Dropout is utilized for regularization, and the final softmax activation produces a probability distribution over the output classes
(one for each activity type). All layers in the neural network utilize the rectified linear unit (ReLU) activation function.

In more detail \ma\ is structured as follows.
The first component of the network is an LSTM (Long Short-Term Memory) layer. LSTMs are a type of recurrent neural network that excels at capturing long-term dependencies in sequential data. This layer has an input size of 4 and a hidden size of 32, which determines the dimensionality of the hidden state. The network is configured with 3 %32 
LSTM layers stacked on top of each other, allowing for the modeling of complex temporal relationships. 
A dropout of 0.1 is applied to the LSTM layer, which helps prevent overfitting by randomly zeroing out elements during training.

The output of the LSTM layer is passed to a fully connected layer with an input size of 32 and an output size of 64. This layer performs a linear transformation of the input data and introduces non-linearity to the network;
A dropout layer with a dropout rate of 0.1 follows the first fully connected layer. Dropout randomly sets a fraction of the input elements to zero during training, providing regularization and reducing the risk of overfitting;
The next fully connected layer takes the output of the previous layer (64-dimensional) and transforms it to a lower-dimensional space of size 32.

Subsequently, another fully connected layer reduces the dimensionality from 32 to 16.
Finally, a softmax activation function is applied to the output of the last fully connected layer. Softmax normalizes the values into a probability distribution, allowing for multiclass classification where each class is assigned a probability.

\paragraph{Tagger model: architecture $\mb$}
The architecture of model $\mb$ only consists of \textsc{dnn} (i.e., feed-forward) layers, which are all assumed to be provided with fixed-length data, based on splitting each trace through a fixed window.
More specifically, a tagger model following this architecture consists of several fully connected layers using the rectified linear unit (ReLU) activation function.

The first layer is a fully connected (linear) layer that takes as input the concatenation of the word embeddings within a window; the size of the input to this layer is determined by $window\_size \cdot 4$, while the output of this layer is a vector of size 128.

The second layer is another fully connected layer that takes the 128-dimensional output of the previous one and project it onto a higher-dimensional space of size 256. 

The third and fourth layers are also fully connected layers that reduce the dimensionality of the input from 256 to 128, and from 128 to 32, respectively.

Finally, the fifth layer is a fully connected layer that maps the 32-dimensional input to the desired output size, i.e. the number of activities; the purpose of this layer is to produce the final features that will be used for the prediction task.

After the last fully connected layer, a softmax function is applied to normalize the output probabilities. This is done to convert the final features into a probability distribution over the possible activities.

\section{Experiments}\label{sec:experiments}

\subsection{Dataset and domain knowledge}
\label{sec:datasets}

The main synthetic dataset (denoted as \textsc{syn}) is inspired by a real-world 
business process, denoted as \textsc{synProc}, of an Italian regional agency. 
\textsc{synProc} involves $16$ high-level activities, \pezzonuovo{named with the capital letters $A, \dots, R$}. 
During the execution of any of these activities, several low-level actions are fired, and each action is recorded in the log as a low-level event. 
Overall, there are $16$ types of (low-level) events (refferred to as $e_1, \dots, e_{16}$) , and each event type
is mapped to up to $5$ different (high-level) activities, that is on average each event type is mapped on $2.7$ activities.
\pezzonuovo{
Specifically, one event type is mapped on five activities, four are mapped on four activities,  four are mapped on three activities,  three are mapped on two activities, and  four are mapped on one activity.
}
\pezzonuovo{The process behavior is described by the must-constraints
$A\Rightarrow_8 C$, 
$O\Rightarrow_6 P$,
$E\Rightarrow_6 F | R$,
$G\Rightarrow_6 I$,
$H\Rightarrow_6 I$, 
$I\Rightarrow_6 D$, the 
not-constraints
$A\Rightarrow_5 \neg B$,
$A\Rightarrow_5 \neg B$,
$A\Rightarrow_1 \neg B$,
$A\Rightarrow_1 \neg B$, 
and the precedence-constraint
$C \Leftarrow_3 D$.}
For each activity $A$, the number of steps $T$ that an instance of $A$ consists of is 
in the range $[1..5]$, 
and the number of instances $X$ of $A$ that can be executed in each process instance 
is in the range $[1..5]$.
\pezzonuovo{In particular, the type-level mapping of \textsc{synProc}
specifies that four activities are mapped on one event, three activities are mapped on two events, five activities are mapped on three events, three activities are mapped on four events, and one activity is mapped on five events.}

Dataset \textsc{syn} was obtained by means of a data generator, 
which takes as input the declarative model of a process $W$ and a mapping between the activities and the events in this model, 
and it returns a set of event traces representing possible instances of $W$.  
Feeding the generator with the process model of \textsc{synProc} and the 
mapping (between event types and activity) described above, we obtained the dataset \textsc{syn}, 
containing $30\,000$ traces, in particular, $10\,000$ traces of length $l$,
for each $l\in \{20,40,60\}$.

Using the same data generator, we also created three additional datasets, referred to as \textsc{syn2}$_{g}$, where $g$ denotes the number of activities associated with each event type by the function cand-act. We used $g \in \{2, 3, 4\}$. Each dataset contains 30,000 traces, specifically 10,000 traces for each length $l \in \{10, 20, 40\}$.

\subsection{Test procedure}\label{sec:predmodels}
The event embeddings were computed by applying Word2Vec model to the event sequences of dataset \textsc{syn} using an output vector size of $4$. 

Independently of the chosen architecture (be it \ma\ or \mb), to learn a trace tagger model, the trace dataset was preliminary split into training and validation sets gathering $80\%$ and $20\%$ of the instances, respectively. 
In all training sessions, we always employed Adam algorithm to optimize the parameters of the tagger models.

In the case of $\ma$ taggers, the training procedure was made span $10$ epochs, using a  learning rate of $10^{-4}$ and a batch size of $32$. 
The architecture of model $\mb$ only consists of \textsc{dnn} (i.e., feed-forward) layers, which are all assumed to be provided with fixed-length trace windows.
Four variants of architecture $\mb$ were tested in the experimentation, which differs in the size $K\in{3,5,7,10}$ of the window employed to eventually split process trace into a vector of length $K$ event embeddings.
These variants are denoted hereinafter as \mb$_K$ with $K\in{3,5,7,10}$.

A fixed setting was used for the arguments $k$ and $\gamma$ of online Algorithm~\ref{algo:analysis}, after providing it with one of these pre-trained taggers and with the AAF-based reasoner, equipped with the specifications of the type-level event-activity mapping and declarative process model.
Specifically, in all the test, we set $k$ equal to the maximum node degree within the bi-partite mapping graph (i.e., the maximum number of activities per event type), and $\gamma=0.001$.
All $\mb_x$ neural networks were trained using the following configuration: the training process spanned 50 epochs, utilizing a batch size of 32 and setting the learning rate to 1e-4.

\subsection{Evaluation metrics and baselines}
\pezzonuovo{We performed experiments to assess both the effectiveness and the efficiency of the proposed approach. 
For the sake of effectiveness evaluation, we measured the prediction accuracy (i.e. the percentage of predictions agreeing with the ground truth), precision, recall and F1 measure, in four usage scenarios:
when using the Tagger only (\accT); when using the Tagger prediction revised by only using the 
function \emph{cand-act} (\accTA); when revising the Tagger prediction by using the Reasoner (\accTR); and
when using the Reasoner only (\accR). 
Clearly enough, the third case corresponds to the approach proposed in this work, while the other three serve as terms of comparison.
These methods will be referred in the following as \emph{Tagger}, \emph{Tagger + Admissibility} and \emph{Tagger + Reasoner} (the proposed approach), and 
 \emph{Reasoner}, respectively ---or more simply as $T$, $T+A$, $T+R$, and $R$, respectively.
It is worth noting that, when evaluating the performance of the Reasoner in terms of accuracy, precision, recall, and F1-score, a methodological challenge arises from the fact that a single event may be associated with multiple candidate activities. As a consequence, these metrics cannot be directly computed in a straightforward manner. To enable a quantitative evaluation, when the Reasoner returns multiple activities for a given event, one activity is randomly selected and treated as the predicted label for the computation of the evaluation metrics.

As for the efficiency, we measured the average time required for yielding a prediction, for each of the first three methods described above: \timeT, \timeTA, and \timeTR. We do not measured the average time required for yielding a prediction for the Reasoner as it coincides with \timeTR\ minus \timeT.

Moreover, to experimentally assess the reduction in information overload achieved by adopting the Tagger+Reasoner approach, as opposed to using the Reasoner alone, we measured the difference between the position of the correct activity in the ranked list produced by the Tagger+Reasoner and the expected position of the correct activity when only the Reasoner is employed.
More precisely, given an event $e_i$, the expected position of the correct activity in the output of the Reasoner, denoted as $EP_{R}(e_i)$, is computed assuming a random ordering of the activities appearing in the interpretations of $e_i$ returned by the Reasoner. Conversely, the position of the correct activity for the same event $e_i$ in the list produced by the Tagger+Reasoner is denoted as $P_{T+R}(e_i)$.
Given a trace $\trace=[e_1,\dots,e_n]$ of length $n$, we define the \emph{reduction of information overload} for $\trace$, denoted as $\Delta_{IO}(\trace)$, as the sum of the differences between these positions over all events in the trace: $\Delta_{IO}(\trace)= \sum_{\i=1}^n EP_{R}(e_i)-P_{T+R}(e_i)$. 
From the analyst's perspective, the reduction of information overload quantifies the number of mapping alternatives that the analyst avoids considering when using the Tagger+Reasoner approach instead of relying solely on the Reasoner.
}

\begin{figure}[htbp]
    \scriptsize
	\begin{tabular}{cccc}
    \multicolumn{4}{c}{\includegraphics[width=0.95\textwidth]{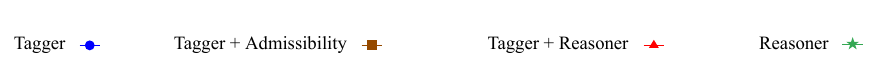}}
        \\
		\includegraphics[width=0.22\textwidth]{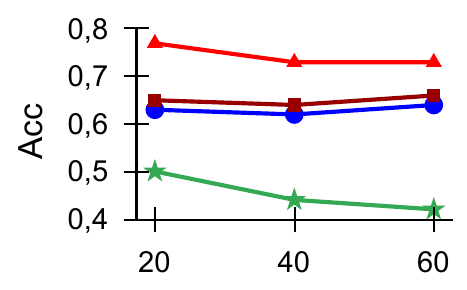} & 
        \includegraphics[width=0.22\textwidth]{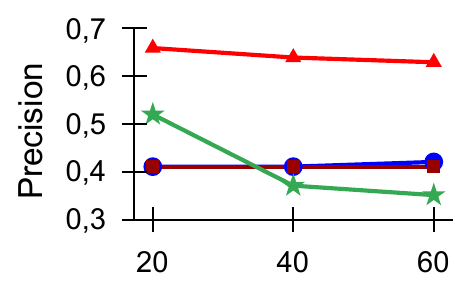} &
        \includegraphics[width=0.22\textwidth]{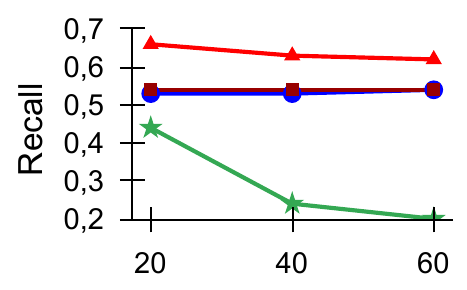} & 
        \includegraphics[width=0.22\textwidth]{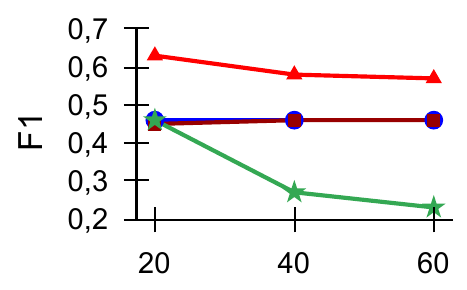} \\
		$(a_1)$ \mb$_3$ Accuracy & $(a_2)$ \mb$_3$ Precision & $(a_3)$ \mb$_3$ Recall & $(a_4)$ \mb$_3$ F1-measure \\[3pt]
		\includegraphics[width=0.22\textwidth]{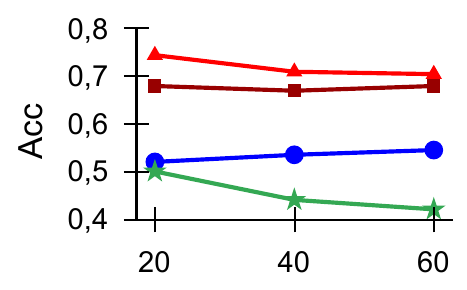} & 
        \includegraphics[width=0.22\textwidth]{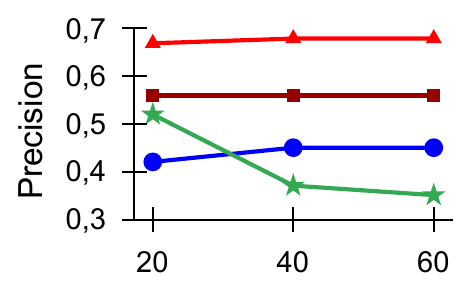} &
        \includegraphics[width=0.22\textwidth]{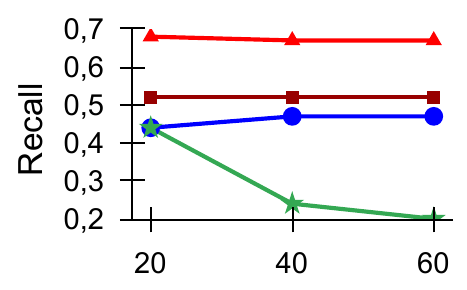} & 
        \includegraphics[width=0.22\textwidth]{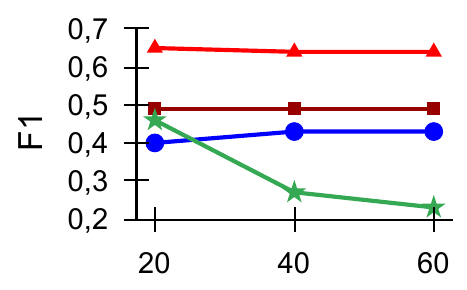} \\
		$(b_1)$ \mb$_5$ Accuracy & $(b_2)$ \mb$_5$ Precision & $(b_3)$ \mb$_5$ Recall & $(b_4)$ \mb$_5$ F1-measure \\[3pt]
        \includegraphics[width=0.22\textwidth]{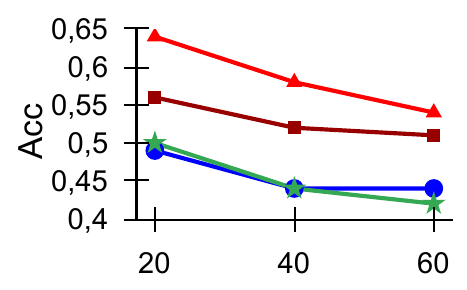} & 
        \includegraphics[width=0.22\textwidth]{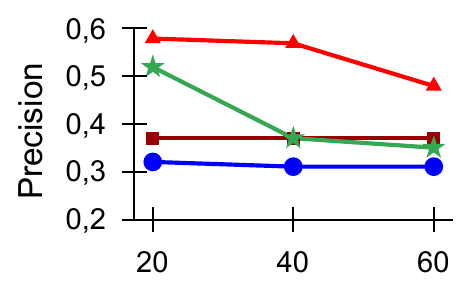} &
        \includegraphics[width=0.22\textwidth]{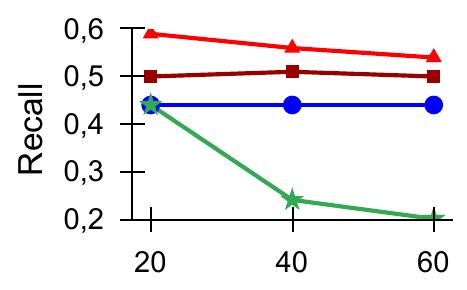} & 
        \includegraphics[width=0.22\textwidth]{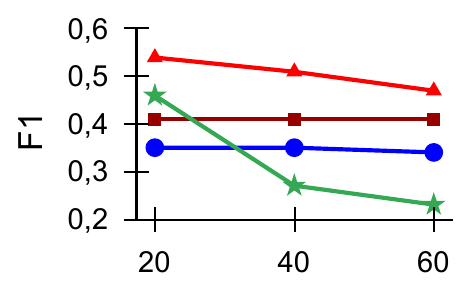} \\
		$(c_1)$ \mb$_7$ Accuracy & $(c_2)$ \mb$_7$ Precision & $(c_3)$ \mb$_7$ Recall & $(c_4)$ \mb$_7$ F1-measure \\[3pt]
        \includegraphics[width=0.22\textwidth]{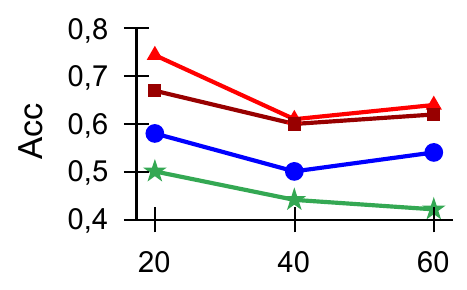} & 
        \includegraphics[width=0.22\textwidth]{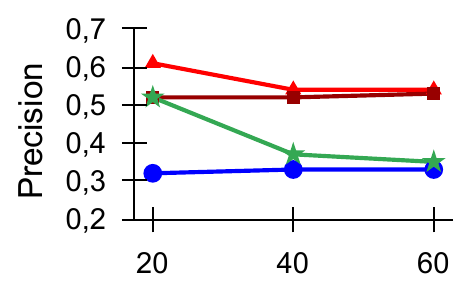} &
        \includegraphics[width=0.22\textwidth]{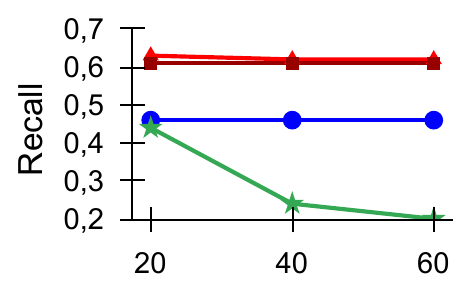} & 
        \includegraphics[width=0.22\textwidth]{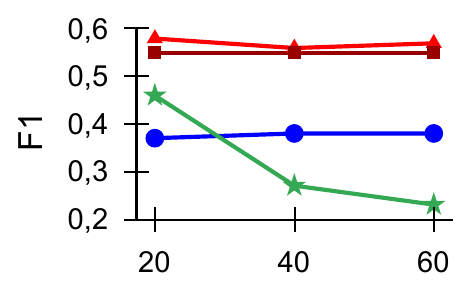} \\
		$(d_1)$ \mb$_{10}$ Accuracy & $(d_2)$ \mb$_{10}$ Precision & $(d_3)$ \mb$_{10}$ Recall & $(d_4)$ \mb$_{10}$ F1-measure \\[3pt]
        \includegraphics[width=0.22\textwidth]{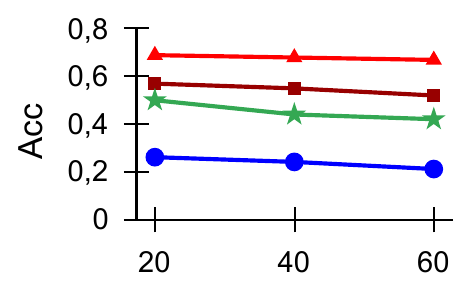} & 
        \includegraphics[width=0.22\textwidth]{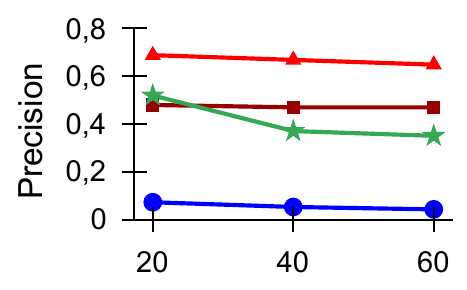} &
        \includegraphics[width=0.22\textwidth]{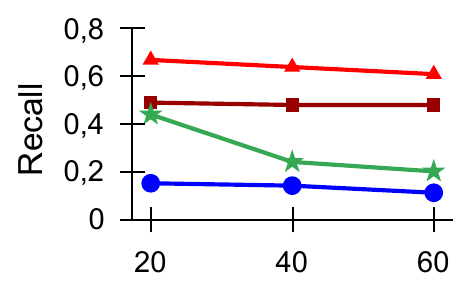} & 
        \includegraphics[width=0.22\textwidth]{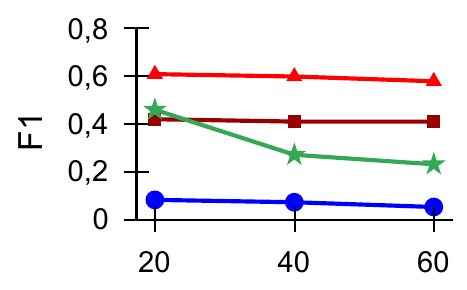} \\
		$(e_1)$ $\ma$ Accuracy & $(e_2)$ $\ma$ Precision & $(e_3)$ $\ma$ Recall & $(e_4)$ $\ma$ F1-measure \\[3pt]
        \multicolumn{4}{c}{\includegraphics[width=0.60\textwidth]{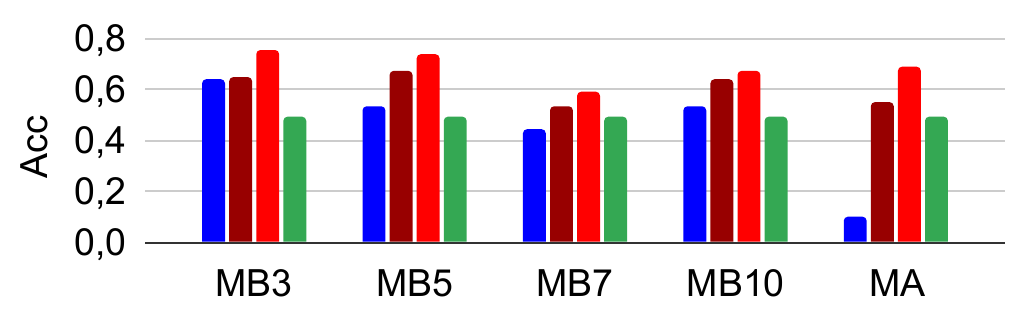}} \\
		\multicolumn{4}{c}{$(f)$}
	\end{tabular}
	\caption{
    Average performance on traces of different lengths (20 to 60) of the methods with neural architectures \mb$_3$ ($a_i$), \mb$_5$ ($b_i$), \mb$_7$ ($c_i$) and \mb$_{10}$ ($d_i$), shown one per row (top to bottom): accuracy ($i=1$), precision ($i=2$), recall ($i=3$) and F1 scores ($i=1$), one per column (left to right).
    The aggregate performances on the whole dataset are reported in $(f)$.}
	\label{fig:accuracy}
\end{figure}

\subsection{Experimental results} \label{sec:exp:res}

The average performance results Figure~\ref{fig:accuracy} reports the averaged performance results (in terms of accuracy, precision, recall and F1) of the four methods (denoted there as Tagger, Tagger+Admissibility,  Tagger+Reasoner, \pezzonuovo{and Reasoner} over \textsc{syn}). 
The letter in the legend identifies the technique and the number the measure, specifically, Figures~\ref{fig:accuracy} show the accuracy in $(a_1-e_1)$, the precision in $(a_2-e_2)$, the recall in $(a_3-e_3)$ and the F1 measure in $(a_4-e_4)$. These are measured w.r.t trace length for the cases that \mb$_3$ $(a_x)$, \mb$_5$ $(b_x)$, \mb$_7$ $(c_x)$, \mb$_{10}$ $(d_x)$, and \ma\ $(e_x)$ are used as tagger, respectively. In Figure~\ref{fig:accuracy}$(f)$ the average measures
for \accT, \accTA\, \accTR\ and \accR\ considering all the trace lengths are reported. 

We first investigate the performance of our approach by comparing it with two settings: predictions obtained using the Tagger alone and predictions obtained using the Tagger augmented with the admissibility test. The diagrams clearly show that \accTR\ consistently outperforms \accT\ across all the considered trace tagger models (Figure~\ref{fig:accuracy}). In particular, highest values for \accTR\ are obtained when \mb$_3$ or \mb$_5$ are used as trace taggers (Figure~\ref{fig:accuracy}($a-b$)). Moreover, in the case that \mb$_5$ is used as tagger model, you can see that \accTR\ is about 20\% higher that \accT. Furthermore, even when the tagger model performs very poorly (see the case of $\ma$ reported in Figure~\ref{fig:accuracy}($f$)) the reasoner can greatly improve its performance (e.g., \accTR\ is much higher than \accT).
Finally, from the results reported in Figure~\ref{fig:accuracy}($g$) it turns out that
the increase in accuracy is much higher when augmenting the tagger models with the reasoner
(\accTR) rater than considering the event-activity mapping only (\accTA).
The findings observed with the accuracy metric are consistently supported by the results obtained from the precision, recall, and F1 measure, confirming the robustness of the evaluation.

\begin{figure}[htbp]
	\includegraphics[width=0.50\textwidth]{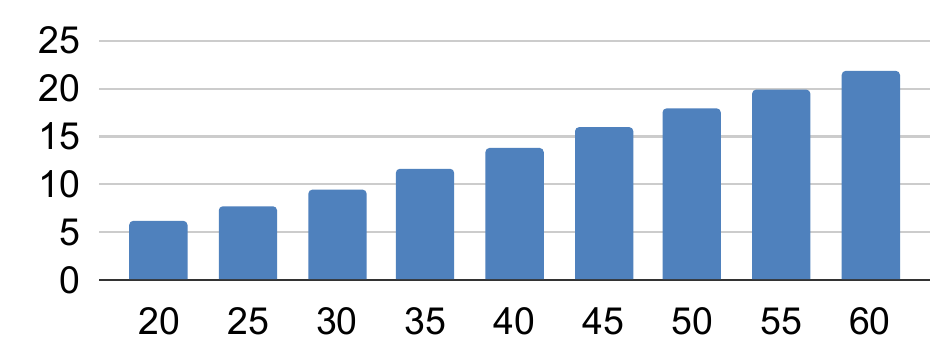}
	\centering
	\caption{\pezzonuovo{Average reduction of information overload $\Delta_{IO}$ for traces of varying lengths in the \textsc{SYN} dataset. For each trace length $x$, the reported reduction was computed by considering only the first $x$ events of traces consisting of $x$ events at least.
    }}
	\label{fig:savings}
\end{figure}

\pezzonuovo{
Regarding the performance of our approach compared with the use of the Reasoner alone, the diagrams indicate that the Tagger+Reasoner consistently outperforms the Reasoner across all evaluated metrics (accuracy, precision, recall, and F1) and for all Tagger configurations that were experimentally assessed.

Moreover, considering the reduction of information overload reported in Figure~\ref{fig:savings}, it can be observed that the number of candidate mapping elements that the analyst can avoid examining by adopting the Tagger+Reasoner approach is substantial (approximately $5$ for traces of length $20$, nearly $15$ for traces of length $40$, and more than $20$ for traces of length $60$), thereby confirming the improvement over the approach presented in~\cite{paper}.

}

\begin{figure}[htbp]
	\begin{tabular}{@{\hspace{-2mm}}c@{\hspace{-12mm}}c}
		\multicolumn{2}{l}{\input{TimeVsLength-legenda.tex}}\\
		\input{TIME-WS3-LENGTH.tex} & \input{TIME-WS5-LENGTH.tex}\\
		$(a)$ & $(b)$ \\
		\input{TIME-WS7-LENGTH.tex} & \input{TIME-WS10-LENGTH.tex}\\
		$(c)$ & $(d)$ \\
		\input{TIME-LSTM-LENGTH.tex} & \input{TIME-ALL-AVGLENGTH.tex}\\
		$(e)$ & $(f)$
	\end{tabular}
	\caption{\timeT\ and \timeTR\ vs trace length for \mb$_3$ $(a)$, 
		\mb$_5$ $(b)$, \mb$_7$ $(c)$, \mb$_{10}$ $(d)$, and \ma ($e$). \timeT, \timeTA, and \timeTR\ over the whole dataset are reported in $(f)$.}
	\label{fig:times}
\end{figure}

\subsection{Results on computation times}
Running times (\timeT\ and \timeTR) are reported in Figures~\ref{fig:times}($a-e$) (they are named Tagger and Tagger+Reasoner, respectively, in the legend keys of Figure~\ref{fig:times}).
\timeTA\ is omitted in Figures~\ref{fig:times}($a-f$) since it is indistinguishable from \timeT. The results shows that \timeTR\ is 3 order of magnitude higher that \timeT\ and \timeTA. However, it is worth noting that \timeTR\ is, on average, around $1$ second, making the proposed approach prone to be effectively used in business process analysis contexts, where the times required to get the analysis results are typically up to tens of seconds.
\pezzonuovo{Furthermore, it is worth pointing out that the execution time of the Tagger+Reasoner approach is nearly identical to that of the Reasoner alone, as the Tagger's computational overhead is negligible compared to the complexity of the reasoning process.}

\begin{figure}[htbp]
		\centering
        \begin{tabular}{c}
            \includegraphics[width=0.95\textwidth]{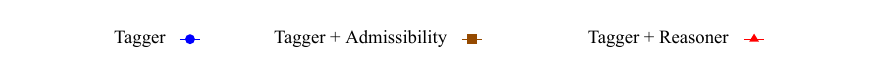} \\
            \includegraphics[width=0.60\textwidth]{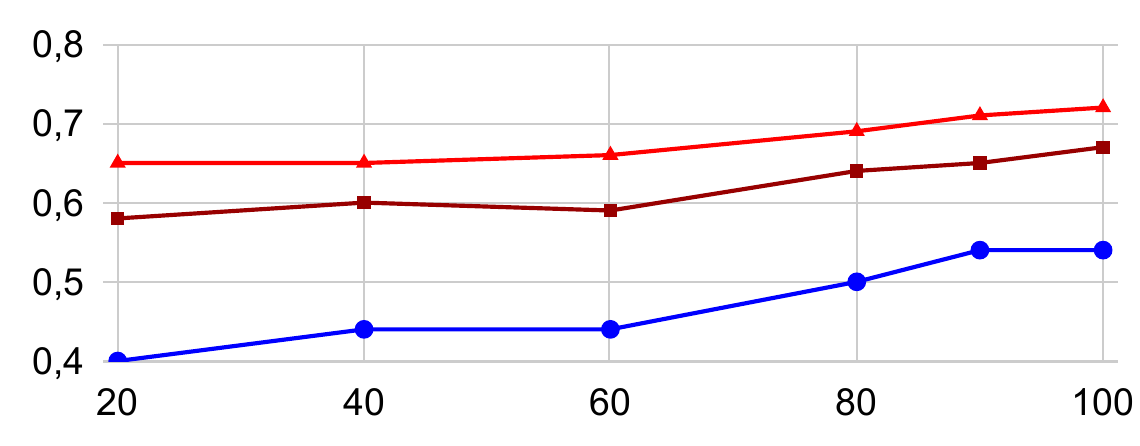}
        \end{tabular}
		\caption{Sensitiveness to the lack of training data: accuracy results obtained when using a tagger model discovered from different random samples of the training set, amounting to different percentages (namely, 20\%, 40\%, 60\%, 80\%, 90\%, and 100\%) of the original training instances.}
		\label{fig:accuracyvsperc}
\end{figure}

\subsection{On the effect of reducing the train set size (Data Efficiency)}
In general, deep models need to be trained with a large number of example data, which may not be available in several real-life contexts, including those where data access restrictions hold and/or human experts can only devote a small amount of time to data annotation~\citep{adadi2021}.

In general, domain knowledge can help compensate for the high risk of making wrong predictions when using a deep model trained over a relatively small amount of example data (and then likely to incur under/over -fitting issues).

To evaluate the advantage of exploiting existing domain knowledge (encoded symbolically in the form of 
inter-activity constraints, and made available to the AAF-based Reasoner), we tested 
the proposed approach (Tagger + Reasoner) in different use cases, corresponding to using a Tagger 
model trained on different amounts of training data.
Specifically, we randomly selected a fraction $r$ of the instances in the training set, by making $r$ vary in $\{20,40,60,80,90,100\}$ (clearly, the extreme case $r=100$ corresponds to using the training set as a whole). 
Then we measured the accuracy score obtained by using the Tagger models discovered from the resulting data samples using the three approaches considered so far: using the Tagger alone (T), after revising the Tagger output with the help of function \emph{cand-act} (Tagger + Admissibility), and using the proposed combination of the Tagger model and revised the Reasoner (Tagger + Reasoner).
\pezzonuovo{In the light of the empirical findings presented in Section \ref{sec:exp:res}, we conducted this and all subsequent experiments using only the $MB_3$ architecture, as it consistently outperformed the other variants of the proposed approach across all the evaluation metrics considered.}

The average results achieved by these three approaches are shown in Figure~\ref{fig:accuracyvsperc} as three different curves, one for each approach.
In general, this figure confirms that the proposed solution (Tagger + Reasoner) neatly outperforms the Tagger model used in isolation, and allows for obtaining better improvements than using only prior knowledge on the event-to-activity mapping. 
Even though the accuracy scores of all the approaches worsen when using smaller training sets, it is important to note that the curve of the proposed method exhibits a far less sharp slope than the other two methods. 
This confirms the benefits of exploiting knowledge about process behaviors in making the effectiveness of the interpretation results more robust to the lack of training data.

\subsection{On the effect of changing the mapping flexibility}

In this section, we present the results of the sensitivity analysis conducted with respect to the number of activities associated with each event type.
To this end, we carried out experiments on the three datasets \textsc{syn2}$_{2}$, \textsc{syn2}$_{3}$, and \textsc{syn2}$_{4}$. For each dataset, we first trained a trace tagger model \textsc{m$_i$} on each dataset and finally executed Algorithm~\ref{algo:analysis} equipped with \textsc{m$_i$} on \textsc{syn2}$_{2}$, for each $i\in \{2,3,4\}$.
The trained trace tagger model is based on the architecture \textsc{mb5} which was the one exhibiting the best performances in the previous experiments.

Analyzing how the effectiveness and computational cost of Algorithm~\ref{algo:analysis} vary across the different datasets provides insight into the impact of the level of uncertainty in the event to activity mapping on the proposed approach.

\begin{figure}[htbp]
	\begin{tabular}{cccc}
        \multicolumn{4}{c}{\includegraphics[width=0.95\textwidth]{legend.pdf}} \\
		\includegraphics[width=0.22\textwidth]{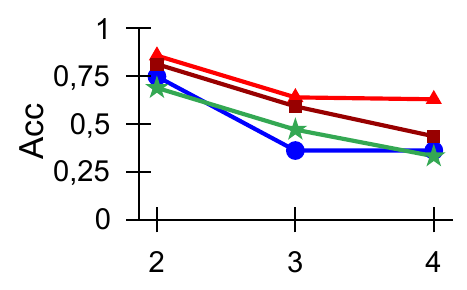} & 
        \includegraphics[width=0.22\textwidth]{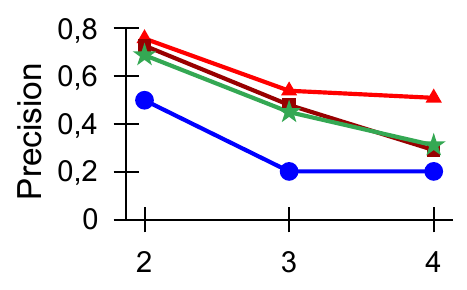} &
        \includegraphics[width=0.22\textwidth]{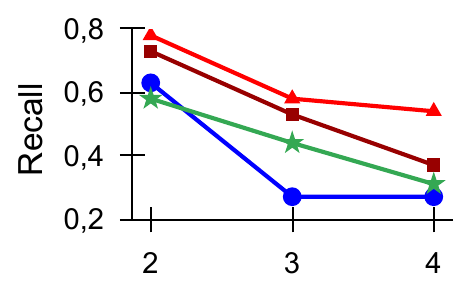} & 
        \includegraphics[width=0.22\textwidth]{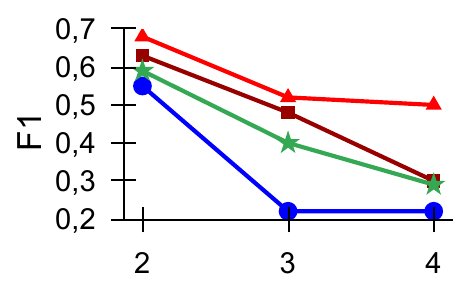} \\ 
        $(a)$ & $(b)$ & $(c)$ & $(d)$ \\
	\end{tabular}
	\caption{Average accuracy \textit{(a)}, precision \textit{(b)}, recall \textit{(c)}, F1 score \textit{(d)} results across different variants of dataset \textsc{Syn2}$_x$, with $x \in \{2,3,4\}$ represented along the horizontal axis.}
	\label{fig:measures_syn2g}
\end{figure}

\begin{figure}[htbp]
	\begin{tabular}{cccc}
        \multicolumn{4}{c}{\includegraphics[width=0.95\textwidth]{legend.pdf}} \\
        \includegraphics[width=0.22\textwidth]{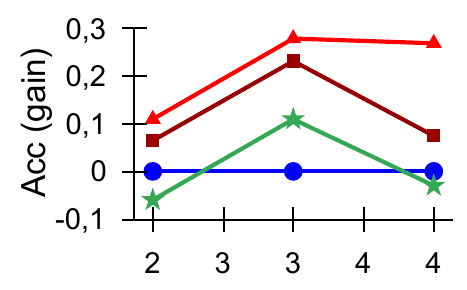} &
        \includegraphics[width=0.22\textwidth]{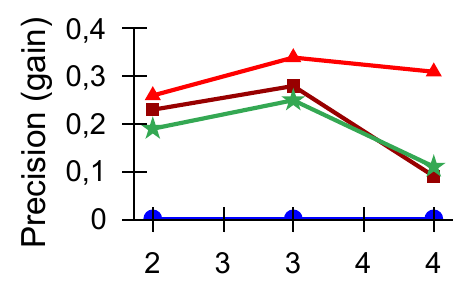} &
        \includegraphics[width=0.22\textwidth]{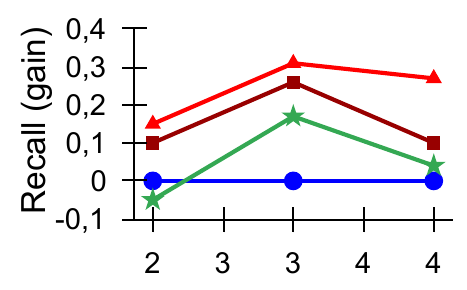} & 
        \includegraphics[width=0.22\textwidth]{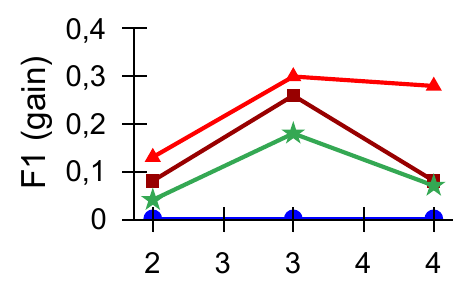} \\
		$(a)$ & $(b)$ & $(c)$ & $(d)$ \\
	\end{tabular}
	\caption{This figure shows the gain in terms of accuracy \textit{(a)}, precision \textit{(b)}, recall \textit{(c)}, F1 score \textit{(d)} across different versions of the \textsc{Syn2}$_x$ dataset, where the variation in $x$ is represented along the x-axis.}
	\label{fig:gains_syn2g}
\end{figure}

\begin{figure}[htbp]
    \begin{tabular}{c}
        \includegraphics[width=0.95\textwidth]{legend_2.pdf} \\
	    \includegraphics[width=0.30\textwidth]{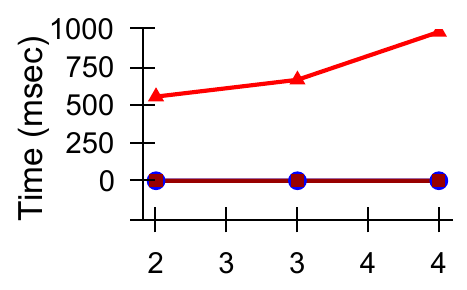}
    \end{tabular}
	\centering
	\caption{This figure shows the execution times across different versions of the \textsc{Syn2}$_x$ dataset, where the variation in $x$ is represented along the x-axis.}
	\label{fig:times_syn2g}
\end{figure}

The results of the experiments are shown in Figure~\ref{fig:measures_syn2g}. The diagrams indicate that as the level of uncertainty in the event-to-activity mapping increases, the effectiveness of the event tagging task decreases (See the diagrams in Figures~\ref{fig:measures_syn2g} $(a), (b), (c)$ and $(d)$) for all methods.
However, the proposed approach (Tagger+Reasoner) manages to achieve the best performances.
\pezzonuovo{
With respect to the comparison with the Reasoner, although the Tagger + Reasoner approach still outperforms the Reasoner alone, the advantages of using Tagger + Reasoner are less pronounced on \textsc{syn2} than in the \textsc{syn} case. A possible explanation is that the Tagger achieves lower accuracy on \textsc{syn2} than on \textsc{syn}.

Looking at the diagrams in Figures~\ref{fig:gains_syn2g} $(a), (b), (c)$ and $(d)$, one can appreciate that the benefits of adopting this approach increase with the level of uncertainty. This confirms that our proposal constitutes a suitable solution in scenarios where simpler methods, such as using the Tagger alone or just corrected with mapping information (Tagger+Admissibility), or the sole Reasoner, cannot achieve a satisfactory level of effectiveness.
}

Finally, Figure~\ref{fig:times_syn2g} shows that although the cost of adopting the proposed approach is significantly higher than that of the other simpler methods, it remains low enough to make its adoption feasible in practice.

\section{Discussion, conclusion and future work}\label{sec:conclusion}

In this work we proposed a novel approach to the problem of interactively interpreting low-level sequence of events, in terms of concepts representing process activities, to enable a business-level monitoring and analysis of process executions. 
This problem has been rephrased into the supervised discovery of neural sequence tagger model, which is then combined with a reasoning module designed to solve the same interpretation problem (as an AAF acceptance problem) based on a given declarative specification of event-activity mappings and of activity-level execution constraints. 
In the resulting neuro-symbolic computation scheme, the candidate interpretations returned by the sequence tagger is refined by reasoner that eventually returns easy-to-interpret even interpretations and explanations. 

This novel approach to event-activity mapping  achieves two main improvements over the state of the art in the field: $1)$ the implicit process knowledge captured, from historical data, by the neural sequence tagger allows for inferring more precise, context-aware, event-activity mappings for any current event $e_{curr}$ and rank them probabilistically, so easing and speeding up the exploration of $e_{curr}$'s interpretations; 
$2)$ the feedback, on the actual validity of the event-to-activity interpretations, provided by  the AAF-based reasoning module, allows for adjusting the tagger predictions and accurately answering a wide range of interpretation queries that the user can pose.

Test results that this hybrid approach is neatly more accurate than a pure ML one (using the neural tagger model alone), especially when using small training sets --see Section~\ref{sec:experiments}.

\paragraph{Novelty and significance} 
The compositional, neuro-symbolic, AI perspective proposed in this work, as a means for improving the effectiveness and efficiency of event interpretation analyses, is definitely novel in the process mining literature.
On the other hand, our research differs from  previous neuro-symbolic approaches (e.g., \cite{DBLP:conf/nesy/AhmedTCBV23}) combining logical constraints and ML models in several key respects: 
\begin{enumerate}
    \item Our behavioral constraints can span over both the input and output variable, rather than over the output only.
    \item The constraints are enforced when using the ML model to estimate the step-wise probabilities of candidate activities, rather than when training the model, in order to both ensure semantically sound query answers and allow the user to change the constraints over time without retraining the ML model.
    \item The user can run  expressive queries that allow her/him turn raw log data into high-level process-aware information.
\end{enumerate}

This original combination of explanaibility, flexibility and efficiency abilities makes the proposed framework a valuable tool for conducting human-in-the-loop, data-driven, process monitoring and analysis in complex real-life engineering applications.

\paragraph{Limitations and future work}
The experiments reported in Section~\ref{sec:experiments} demonstrate that the proposed approach significantly enhances the effectiveness of a trace tagging model, particularly in scenarios where the available (annotated) example traces are insufficient to train a high-quality model. However, in certain (extreme) application contexts, the scarcity of annotated traces can be so severe that even the proposed method fails to achieve satisfactory performance.
In fact, in some cases, annotating the steps of historical log traces with corresponding ground-truth activities is a time-consuming and complex manual task that requires both expertise and effort. As a result, the number of annotated traces available may be insufficient to train a high-quality tagging model, even when using the proposed framework.
To properly address these situations, as future work, we plan to investigate the following 
strategies for improving the training of any trace tagger model $M$ (as a complement to amending its output with the help of a symbolic reasoner):
\begin{itemize}
\item 
Extending the training procedure in a way that a (usually wider and easier to obtain) collection of 
unlabeled log traces of the form $U = \{ \trace \} \subset \U^*$ can be used in addition to the annotated 
traces $L$. To get some form of supervision from the examples in $U$, an entropy-regularization 
loss term \cite{entropy-regularization} can be computed over each of these examples, which penalizes cases 
where the predicted activity-wise probability distribution is not skewed (the higher the entropy of 
the predicted distribution, the higher the loss).
\item 
After inducing a preliminary version of $M$, one can think of exploiting an Active Learning \cite{AL} scheme to assist the expert in assigning ground-truth activity labels to a small number, say $q$, of unlabelled traces (which are expected to help improve $M$, once exploited as additional training examples).
In particular, we will investigate using two ranking criteria to automatically select useful traces among the unlabelled ones, based on the current version $M$ of the tagger model: 
$(i)$ ranking the traces on the basis of the average/maximum entropy of their per-step predictions;
$(ii)$ ranking the traces on the basis of the expected number of times their per-step predictions are 
deemed invalid by the reasoner $R$.
\end{itemize}

The main limitation to the effectiveness of the proposed approach lies in the absence (or scarcity) of domain knowledge, particularly in cases where the known behavioral constraints are too few to effectively guide the predictions of the tagger model.
Nonetheless, in certain cases, even when explicit knowledge of the behavioral constraints that hold in the domain is scarce, it can be enhanced by incorporating alternative forms of knowledge related to typical process executions. For instance, probabilistic information such as ``In 80\% of cases, an execution of activity A immediately follows an execution of activity B" can serve as a valuable supplement.

Unfortunately, the proposed AAF-based approach is unable to explicitly exploit such knowledge. It can only leverage this type of information if the tagger model has implicitly learned it during the training phase, as there are no mechanisms for incorporating it directly.
The design of such a mechanism is complex and will be addressed in future work.

\subsection*{Declarations}
\noindent \textbf{Funding} The authors received no specific funding for this study.\\
\textbf{Conflict of interest/Competing interests} The authors declare no conflict of interest.\\
\noindent 
\textbf{Ethics approval} Not Applicable.\\ 
\noindent 
\textbf{Data availability}
Data contain private information that cannot be disclosed.\\
\noindent 
\textbf{Authors' contribution}
All authors have contributed equally to this work.\\
\thispagestyle{empty}
\noindent 
\textbf{Acknowledgement}
This work was partly supported by project FAIR - Future AI Research - Spoke 9 (Directorial Decree no. 1243, August 2nd, 2022; PE 0000013; CUP B53C22003630006), under the NRRP (National Recovery and Resilience Plan) MUR program (Mission 4, Component 2 Investment 1.3) funded by the European Union - NextGenerationEU.

\bibliography{bibliography}% common bib file

\end{document}

%% file: TimeVsLength-legenda.tex
% GNUPLOT: LaTeX picture with Postscript
\begingroup
  % Encoding inside the plot.  In the header of your document, this encoding
  % should to defined, e.g., by using
  % \usepackage[cp1252,<other encodings>]{inputenc}
  \selectfont
  \makeatletter
  \providecommand\color[2][]{%
    \GenericError{(gnuplot) \space\space\space\@spaces}{%
      Package color not loaded in conjunction with
      terminal option `colourtext'%
    }{See the gnuplot documentation for explanation.%
    }{Either use 'blacktext' in gnuplot or load the package
      color.sty in LaTeX.}%
    \renewcommand\color[2][]{}%
  }%
  \providecommand\includegraphics[2][]{%
    \GenericError{(gnuplot) \space\space\space\@spaces}{%
      Package graphicx or graphics not loaded%
    }{See the gnuplot documentation for explanation.%
    }{The gnuplot epslatex terminal needs graphicx.sty or graphics.sty.}%
    \renewcommand\includegraphics[2][]{}%
  }%
  \providecommand\rotatebox[2]{#2}%
  \@ifundefined{ifGPcolor}{%
    \newif\ifGPcolor
    \GPcolorfalse
  }{}%
  \@ifundefined{ifGPblacktext}{%
    \newif\ifGPblacktext
    \GPblacktexttrue
  }{}%
  % define a \g@addto@macro without @ in the name:
  \let\gplgaddtomacro\g@addto@macro
  % define empty templates for all commands taking text:
  \gdef\gplbacktext{}%
  \gdef\gplfronttext{}%
  \makeatother
  \ifGPblacktext
    % no textcolor at all
    \def\colorrgb#1{}%
    \def\colorgray#1{}%
  \else
    % gray or color?
    \ifGPcolor
      \def\colorrgb#1{\color[rgb]{#1}}%
      \def\colorgray#1{\color[gray]{#1}}%
      \expandafter\def\csname LTw\endcsname{\color{white}}%
      \expandafter\def\csname LTb\endcsname{\color{black}}%
      \expandafter\def\csname LTa\endcsname{\color{black}}%
      \expandafter\def\csname LT0\endcsname{\color[rgb]{1,0,0}}%
      \expandafter\def\csname LT1\endcsname{\color[rgb]{0,1,0}}%
      \expandafter\def\csname LT2\endcsname{\color[rgb]{0,0,1}}%
      \expandafter\def\csname LT3\endcsname{\color[rgb]{1,0,1}}%
      \expandafter\def\csname LT4\endcsname{\color[rgb]{0,1,1}}%
      \expandafter\def\csname LT5\endcsname{\color[rgb]{1,1,0}}%
      \expandafter\def\csname LT6\endcsname{\color[rgb]{0,0,0}}%
      \expandafter\def\csname LT7\endcsname{\color[rgb]{1,0.3,0}}%
      \expandafter\def\csname LT8\endcsname{\color[rgb]{0.5,0.5,0.5}}%
    \else
      % gray
      \def\colorrgb#1{\color{black}}%
      \def\colorgray#1{\color[gray]{#1}}%
      \expandafter\def\csname LTw\endcsname{\color{white}}%
      \expandafter\def\csname LTb\endcsname{\color{black}}%
      \expandafter\def\csname LTa\endcsname{\color{black}}%
      \expandafter\def\csname LT0\endcsname{\color{black}}%
      \expandafter\def\csname LT1\endcsname{\color{black}}%
      \expandafter\def\csname LT2\endcsname{\color{black}}%
      \expandafter\def\csname LT3\endcsname{\color{black}}%
      \expandafter\def\csname LT4\endcsname{\color{black}}%
      \expandafter\def\csname LT5\endcsname{\color{black}}%
      \expandafter\def\csname LT6\endcsname{\color{black}}%
      \expandafter\def\csname LT7\endcsname{\color{black}}%
      \expandafter\def\csname LT8\endcsname{\color{black}}%
    \fi
  \fi
    \setlength{\unitlength}{0.0500bp}%
    \ifx\gptboxheight\undefined%
      \newlength{\gptboxheight}%
      \newlength{\gptboxwidth}%
      \newsavebox{\gptboxtext}%
    \fi%
    \setlength{\fboxrule}{0.5pt}%
    \setlength{\fboxsep}{1pt}%
    \definecolor{tbcol}{rgb}{1,1,1}%
\begin{picture}(7776.00,504.00)%
    \gplgaddtomacro\gplbacktext{%
    }%
    \gplgaddtomacro\gplfronttext{%
      \csname LTb\endcsname%%
      \put(2706,267){\makebox(0,0)[r]{\strut{}Tagger}}%
      \csname LTb\endcsname%%
      \put(5409,267){\makebox(0,0)[r]{\strut{}Tagger + Reasoner}}%
    }%
    \gplbacktext
    \put(0,0){\includegraphics[width={388.80bp},height={25.20bp}]{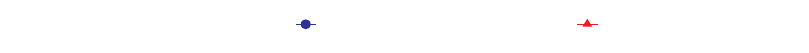}}%
    \gplfronttext
  \end{picture}%
\endgroup

%% file: TIME-WS3-LENGTH.tex
% GNUPLOT: LaTeX picture with Postscript
\begingroup
  % Encoding inside the plot.  In the header of your document, this encoding
  % should to defined, e.g., by using
  % \usepackage[cp1252,<other encodings>]{inputenc}
  \selectfont
  \makeatletter
  \providecommand\color[2][]{%
    \GenericError{(gnuplot) \space\space\space\@spaces}{%
      Package color not loaded in conjunction with
      terminal option `colourtext'%
    }{See the gnuplot documentation for explanation.%
    }{Either use 'blacktext' in gnuplot or load the package
      color.sty in LaTeX.}%
    \renewcommand\color[2][]{}%
  }%
  \providecommand\includegraphics[2][]{%
    \GenericError{(gnuplot) \space\space\space\@spaces}{%
      Package graphicx or graphics not loaded%
    }{See the gnuplot documentation for explanation.%
    }{The gnuplot epslatex terminal needs graphicx.sty or graphics.sty.}%
    \renewcommand\includegraphics[2][]{}%
  }%
  \providecommand\rotatebox[2]{#2}%
  \@ifundefined{ifGPcolor}{%
    \newif\ifGPcolor
    \GPcolorfalse
  }{}%
  \@ifundefined{ifGPblacktext}{%
    \newif\ifGPblacktext
    \GPblacktexttrue
  }{}%
  % define a \g@addto@macro without @ in the name:
  \let\gplgaddtomacro\g@addto@macro
  % define empty templates for all commands taking text:
  \gdef\gplbacktext{}%
  \gdef\gplfronttext{}%
  \makeatother
  \ifGPblacktext
    % no textcolor at all
    \def\colorrgb#1{}%
    \def\colorgray#1{}%
  \else
    % gray or color?
    \ifGPcolor
      \def\colorrgb#1{\color[rgb]{#1}}%
      \def\colorgray#1{\color[gray]{#1}}%
      \expandafter\def\csname LTw\endcsname{\color{white}}%
      \expandafter\def\csname LTb\endcsname{\color{black}}%
      \expandafter\def\csname LTa\endcsname{\color{black}}%
      \expandafter\def\csname LT0\endcsname{\color[rgb]{1,0,0}}%
      \expandafter\def\csname LT1\endcsname{\color[rgb]{0,1,0}}%
      \expandafter\def\csname LT2\endcsname{\color[rgb]{0,0,1}}%
      \expandafter\def\csname LT3\endcsname{\color[rgb]{1,0,1}}%
      \expandafter\def\csname LT4\endcsname{\color[rgb]{0,1,1}}%
      \expandafter\def\csname LT5\endcsname{\color[rgb]{1,1,0}}%
      \expandafter\def\csname LT6\endcsname{\color[rgb]{0,0,0}}%
      \expandafter\def\csname LT7\endcsname{\color[rgb]{1,0.3,0}}%
      \expandafter\def\csname LT8\endcsname{\color[rgb]{0.5,0.5,0.5}}%
    \else
      % gray
      \def\colorrgb#1{\color{black}}%
      \def\colorgray#1{\color[gray]{#1}}%
      \expandafter\def\csname LTw\endcsname{\color{white}}%
      \expandafter\def\csname LTb\endcsname{\color{black}}%
      \expandafter\def\csname LTa\endcsname{\color{black}}%
      \expandafter\def\csname LT0\endcsname{\color{black}}%
      \expandafter\def\csname LT1\endcsname{\color{black}}%
      \expandafter\def\csname LT2\endcsname{\color{black}}%
      \expandafter\def\csname LT3\endcsname{\color{black}}%
      \expandafter\def\csname LT4\endcsname{\color{black}}%
      \expandafter\def\csname LT5\endcsname{\color{black}}%
      \expandafter\def\csname LT6\endcsname{\color{black}}%
      \expandafter\def\csname LT7\endcsname{\color{black}}%
      \expandafter\def\csname LT8\endcsname{\color{black}}%
    \fi
  \fi
    \setlength{\unitlength}{0.0500bp}%
    \ifx\gptboxheight\undefined%
      \newlength{\gptboxheight}%
      \newlength{\gptboxwidth}%
      \newsavebox{\gptboxtext}%
    \fi%
    \setlength{\fboxrule}{0.5pt}%
    \setlength{\fboxsep}{1pt}%
    \definecolor{tbcol}{rgb}{1,1,1}%
\begin{picture}(3168.00,2016.00)%
    \gplgaddtomacro\gplbacktext{%
      \csname LTb\endcsname%%
      \put(924,704){\makebox(0,0)[r]{\strut{}\small$0.1$}}%
      \put(924,936){\makebox(0,0)[r]{\strut{}\small$1$}}%
      \put(924,1169){\makebox(0,0)[r]{\strut{}\small$10$}}%
      \put(924,1401){\makebox(0,0)[r]{\strut{}\small$100$}}%
      \put(924,1634){\makebox(0,0)[r]{\strut{}\small$1000$}}%
      \put(924,1796){\makebox(0,0)[r]{\strut{}\small$5000$}}%
      \put(990,484){\makebox(0,0){\strut{}\small 20}}%
      \put(1881,484){\makebox(0,0){\strut{}\small 40}}%
      \put(2771,484){\makebox(0,0){\strut{}\small 60}}%
      \put(322,634){\rotatebox{90}{\makebox(0,0)[l]{\strut{}Time (msec)}}}%
    }%
    \gplgaddtomacro\gplfronttext{%
      \csname LTb\endcsname%%
      \put(1880,154){\makebox(0,0){\strut{}\small Trace length}}%
    }%
    \gplbacktext
    \put(0,0){\includegraphics[width={158.40bp},height={100.80bp}]{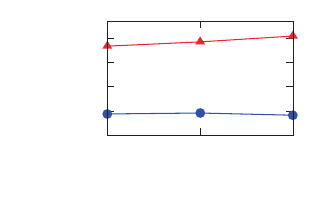}}%
    \gplfronttext
  \end{picture}%
\endgroup

%% file: TIME-WS5-LENGTH.tex
% GNUPLOT: LaTeX picture with Postscript
\begingroup
  % Encoding inside the plot.  In the header of your document, this encoding
  % should to defined, e.g., by using
  % \usepackage[cp1252,<other encodings>]{inputenc}
  \selectfont
  \makeatletter
  \providecommand\color[2][]{%
    \GenericError{(gnuplot) \space\space\space\@spaces}{%
      Package color not loaded in conjunction with
      terminal option `colourtext'%
    }{See the gnuplot documentation for explanation.%
    }{Either use 'blacktext' in gnuplot or load the package
      color.sty in LaTeX.}%
    \renewcommand\color[2][]{}%
  }%
  \providecommand\includegraphics[2][]{%
    \GenericError{(gnuplot) \space\space\space\@spaces}{%
      Package graphicx or graphics not loaded%
    }{See the gnuplot documentation for explanation.%
    }{The gnuplot epslatex terminal needs graphicx.sty or graphics.sty.}%
    \renewcommand\includegraphics[2][]{}%
  }%
  \providecommand\rotatebox[2]{#2}%
  \@ifundefined{ifGPcolor}{%
    \newif\ifGPcolor
    \GPcolorfalse
  }{}%
  \@ifundefined{ifGPblacktext}{%
    \newif\ifGPblacktext
    \GPblacktexttrue
  }{}%
  % define a \g@addto@macro without @ in the name:
  \let\gplgaddtomacro\g@addto@macro
  % define empty templates for all commands taking text:
  \gdef\gplbacktext{}%
  \gdef\gplfronttext{}%
  \makeatother
  \ifGPblacktext
    % no textcolor at all
    \def\colorrgb#1{}%
    \def\colorgray#1{}%
  \else
    % gray or color?
    \ifGPcolor
      \def\colorrgb#1{\color[rgb]{#1}}%
      \def\colorgray#1{\color[gray]{#1}}%
      \expandafter\def\csname LTw\endcsname{\color{white}}%
      \expandafter\def\csname LTb\endcsname{\color{black}}%
      \expandafter\def\csname LTa\endcsname{\color{black}}%
      \expandafter\def\csname LT0\endcsname{\color[rgb]{1,0,0}}%
      \expandafter\def\csname LT1\endcsname{\color[rgb]{0,1,0}}%
      \expandafter\def\csname LT2\endcsname{\color[rgb]{0,0,1}}%
      \expandafter\def\csname LT3\endcsname{\color[rgb]{1,0,1}}%
      \expandafter\def\csname LT4\endcsname{\color[rgb]{0,1,1}}%
      \expandafter\def\csname LT5\endcsname{\color[rgb]{1,1,0}}%
      \expandafter\def\csname LT6\endcsname{\color[rgb]{0,0,0}}%
      \expandafter\def\csname LT7\endcsname{\color[rgb]{1,0.3,0}}%
      \expandafter\def\csname LT8\endcsname{\color[rgb]{0.5,0.5,0.5}}%
    \else
      % gray
      \def\colorrgb#1{\color{black}}%
      \def\colorgray#1{\color[gray]{#1}}%
      \expandafter\def\csname LTw\endcsname{\color{white}}%
      \expandafter\def\csname LTb\endcsname{\color{black}}%
      \expandafter\def\csname LTa\endcsname{\color{black}}%
      \expandafter\def\csname LT0\endcsname{\color{black}}%
      \expandafter\def\csname LT1\endcsname{\color{black}}%
      \expandafter\def\csname LT2\endcsname{\color{black}}%
      \expandafter\def\csname LT3\endcsname{\color{black}}%
      \expandafter\def\csname LT4\endcsname{\color{black}}%
      \expandafter\def\csname LT5\endcsname{\color{black}}%
      \expandafter\def\csname LT6\endcsname{\color{black}}%
      \expandafter\def\csname LT7\endcsname{\color{black}}%
      \expandafter\def\csname LT8\endcsname{\color{black}}%
    \fi
  \fi
    \setlength{\unitlength}{0.0500bp}%
    \ifx\gptboxheight\undefined%
      \newlength{\gptboxheight}%
      \newlength{\gptboxwidth}%
      \newsavebox{\gptboxtext}%
    \fi%
    \setlength{\fboxrule}{0.5pt}%
    \setlength{\fboxsep}{1pt}%
    \definecolor{tbcol}{rgb}{1,1,1}%
\begin{picture}(3168.00,2016.00)%
    \gplgaddtomacro\gplbacktext{%
      \csname LTb\endcsname%%
      \put(924,704){\makebox(0,0)[r]{\strut{}\small$0.1$}}%
      \put(924,936){\makebox(0,0)[r]{\strut{}\small$1$}}%
      \put(924,1169){\makebox(0,0)[r]{\strut{}\small$10$}}%
      \put(924,1401){\makebox(0,0)[r]{\strut{}\small$100$}}%
      \put(924,1634){\makebox(0,0)[r]{\strut{}\small$1000$}}%
      \put(924,1796){\makebox(0,0)[r]{\strut{}\small$5000$}}%
      \put(990,484){\makebox(0,0){\strut{}\small 20}}%
      \put(1881,484){\makebox(0,0){\strut{}\small 40}}%
      \put(2771,484){\makebox(0,0){\strut{}\small 60}}%
      \put(322,634){\rotatebox{90}{\makebox(0,0)[l]{\strut{}Time (msec)}}}%
    }%
    \gplgaddtomacro\gplfronttext{%
      \csname LTb\endcsname%%
      \put(1880,154){\makebox(0,0){\strut{}\small Trace length}}%
    }%
    \gplbacktext
    \put(0,0){\includegraphics[width={158.40bp},height={100.80bp}]{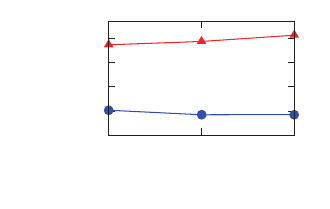}}%
    \gplfronttext
  \end{picture}%
\endgroup

%% file: TIME-WS7-LENGTH.tex
% GNUPLOT: LaTeX picture with Postscript
\begingroup
  % Encoding inside the plot.  In the header of your document, this encoding
  % should to defined, e.g., by using
  % \usepackage[cp1252,<other encodings>]{inputenc}
  \selectfont
  \makeatletter
  \providecommand\color[2][]{%
    \GenericError{(gnuplot) \space\space\space\@spaces}{%
      Package color not loaded in conjunction with
      terminal option `colourtext'%
    }{See the gnuplot documentation for explanation.%
    }{Either use 'blacktext' in gnuplot or load the package
      color.sty in LaTeX.}%
    \renewcommand\color[2][]{}%
  }%
  \providecommand\includegraphics[2][]{%
    \GenericError{(gnuplot) \space\space\space\@spaces}{%
      Package graphicx or graphics not loaded%
    }{See the gnuplot documentation for explanation.%
    }{The gnuplot epslatex terminal needs graphicx.sty or graphics.sty.}%
    \renewcommand\includegraphics[2][]{}%
  }%
  \providecommand\rotatebox[2]{#2}%
  \@ifundefined{ifGPcolor}{%
    \newif\ifGPcolor
    \GPcolorfalse
  }{}%
  \@ifundefined{ifGPblacktext}{%
    \newif\ifGPblacktext
    \GPblacktexttrue
  }{}%
  % define a \g@addto@macro without @ in the name:
  \let\gplgaddtomacro\g@addto@macro
  % define empty templates for all commands taking text:
  \gdef\gplbacktext{}%
  \gdef\gplfronttext{}%
  \makeatother
  \ifGPblacktext
    % no textcolor at all
    \def\colorrgb#1{}%
    \def\colorgray#1{}%
  \else
    % gray or color?
    \ifGPcolor
      \def\colorrgb#1{\color[rgb]{#1}}%
      \def\colorgray#1{\color[gray]{#1}}%
      \expandafter\def\csname LTw\endcsname{\color{white}}%
      \expandafter\def\csname LTb\endcsname{\color{black}}%
      \expandafter\def\csname LTa\endcsname{\color{black}}%
      \expandafter\def\csname LT0\endcsname{\color[rgb]{1,0,0}}%
      \expandafter\def\csname LT1\endcsname{\color[rgb]{0,1,0}}%
      \expandafter\def\csname LT2\endcsname{\color[rgb]{0,0,1}}%
      \expandafter\def\csname LT3\endcsname{\color[rgb]{1,0,1}}%
      \expandafter\def\csname LT4\endcsname{\color[rgb]{0,1,1}}%
      \expandafter\def\csname LT5\endcsname{\color[rgb]{1,1,0}}%
      \expandafter\def\csname LT6\endcsname{\color[rgb]{0,0,0}}%
      \expandafter\def\csname LT7\endcsname{\color[rgb]{1,0.3,0}}%
      \expandafter\def\csname LT8\endcsname{\color[rgb]{0.5,0.5,0.5}}%
    \else
      % gray
      \def\colorrgb#1{\color{black}}%
      \def\colorgray#1{\color[gray]{#1}}%
      \expandafter\def\csname LTw\endcsname{\color{white}}%
      \expandafter\def\csname LTb\endcsname{\color{black}}%
      \expandafter\def\csname LTa\endcsname{\color{black}}%
      \expandafter\def\csname LT0\endcsname{\color{black}}%
      \expandafter\def\csname LT1\endcsname{\color{black}}%
      \expandafter\def\csname LT2\endcsname{\color{black}}%
      \expandafter\def\csname LT3\endcsname{\color{black}}%
      \expandafter\def\csname LT4\endcsname{\color{black}}%
      \expandafter\def\csname LT5\endcsname{\color{black}}%
      \expandafter\def\csname LT6\endcsname{\color{black}}%
      \expandafter\def\csname LT7\endcsname{\color{black}}%
      \expandafter\def\csname LT8\endcsname{\color{black}}%
    \fi
  \fi
    \setlength{\unitlength}{0.0500bp}%
    \ifx\gptboxheight\undefined%
      \newlength{\gptboxheight}%
      \newlength{\gptboxwidth}%
      \newsavebox{\gptboxtext}%
    \fi%
    \setlength{\fboxrule}{0.5pt}%
    \setlength{\fboxsep}{1pt}%
    \definecolor{tbcol}{rgb}{1,1,1}%
\begin{picture}(3168.00,2016.00)%
    \gplgaddtomacro\gplbacktext{%
      \csname LTb\endcsname%%
      \put(924,704){\makebox(0,0)[r]{\strut{}\small$0.1$}}%
      \put(924,936){\makebox(0,0)[r]{\strut{}\small$1$}}%
      \put(924,1169){\makebox(0,0)[r]{\strut{}\small$10$}}%
      \put(924,1401){\makebox(0,0)[r]{\strut{}\small$100$}}%
      \put(924,1634){\makebox(0,0)[r]{\strut{}\small$1000$}}%
      \put(924,1796){\makebox(0,0)[r]{\strut{}\small$5000$}}%
      \put(990,484){\makebox(0,0){\strut{}\small 20}}%
      \put(1881,484){\makebox(0,0){\strut{}\small 40}}%
      \put(2771,484){\makebox(0,0){\strut{}\small 60}}%
      \put(322,634){\rotatebox{90}{\makebox(0,0)[l]{\strut{}Time (msec)}}}%
    }%
    \gplgaddtomacro\gplfronttext{%
      \csname LTb\endcsname%%
      \put(1880,154){\makebox(0,0){\strut{}\small Trace length}}%
    }%
    \gplbacktext
    \put(0,0){\includegraphics[width={158.40bp},height={100.80bp}]{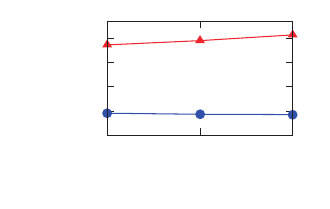}}%
    \gplfronttext
  \end{picture}%
\endgroup

%% file: TIME-WS10-LENGTH.tex
% GNUPLOT: LaTeX picture with Postscript
\begingroup
  % Encoding inside the plot.  In the header of your document, this encoding
  % should to defined, e.g., by using
  % \usepackage[cp1252,<other encodings>]{inputenc}
  \selectfont
  \makeatletter
  \providecommand\color[2][]{%
    \GenericError{(gnuplot) \space\space\space\@spaces}{%
      Package color not loaded in conjunction with
      terminal option `colourtext'%
    }{See the gnuplot documentation for explanation.%
    }{Either use 'blacktext' in gnuplot or load the package
      color.sty in LaTeX.}%
    \renewcommand\color[2][]{}%
  }%
  \providecommand\includegraphics[2][]{%
    \GenericError{(gnuplot) \space\space\space\@spaces}{%
      Package graphicx or graphics not loaded%
    }{See the gnuplot documentation for explanation.%
    }{The gnuplot epslatex terminal needs graphicx.sty or graphics.sty.}%
    \renewcommand\includegraphics[2][]{}%
  }%
  \providecommand\rotatebox[2]{#2}%
  \@ifundefined{ifGPcolor}{%
    \newif\ifGPcolor
    \GPcolorfalse
  }{}%
  \@ifundefined{ifGPblacktext}{%
    \newif\ifGPblacktext
    \GPblacktexttrue
  }{}%
  % define a \g@addto@macro without @ in the name:
  \let\gplgaddtomacro\g@addto@macro
  % define empty templates for all commands taking text:
  \gdef\gplbacktext{}%
  \gdef\gplfronttext{}%
  \makeatother
  \ifGPblacktext
    % no textcolor at all
    \def\colorrgb#1{}%
    \def\colorgray#1{}%
  \else
    % gray or color?
    \ifGPcolor
      \def\colorrgb#1{\color[rgb]{#1}}%
      \def\colorgray#1{\color[gray]{#1}}%
      \expandafter\def\csname LTw\endcsname{\color{white}}%
      \expandafter\def\csname LTb\endcsname{\color{black}}%
      \expandafter\def\csname LTa\endcsname{\color{black}}%
      \expandafter\def\csname LT0\endcsname{\color[rgb]{1,0,0}}%
      \expandafter\def\csname LT1\endcsname{\color[rgb]{0,1,0}}%
      \expandafter\def\csname LT2\endcsname{\color[rgb]{0,0,1}}%
      \expandafter\def\csname LT3\endcsname{\color[rgb]{1,0,1}}%
      \expandafter\def\csname LT4\endcsname{\color[rgb]{0,1,1}}%
      \expandafter\def\csname LT5\endcsname{\color[rgb]{1,1,0}}%
      \expandafter\def\csname LT6\endcsname{\color[rgb]{0,0,0}}%
      \expandafter\def\csname LT7\endcsname{\color[rgb]{1,0.3,0}}%
      \expandafter\def\csname LT8\endcsname{\color[rgb]{0.5,0.5,0.5}}%
    \else
      % gray
      \def\colorrgb#1{\color{black}}%
      \def\colorgray#1{\color[gray]{#1}}%
      \expandafter\def\csname LTw\endcsname{\color{white}}%
      \expandafter\def\csname LTb\endcsname{\color{black}}%
      \expandafter\def\csname LTa\endcsname{\color{black}}%
      \expandafter\def\csname LT0\endcsname{\color{black}}%
      \expandafter\def\csname LT1\endcsname{\color{black}}%
      \expandafter\def\csname LT2\endcsname{\color{black}}%
      \expandafter\def\csname LT3\endcsname{\color{black}}%
      \expandafter\def\csname LT4\endcsname{\color{black}}%
      \expandafter\def\csname LT5\endcsname{\color{black}}%
      \expandafter\def\csname LT6\endcsname{\color{black}}%
      \expandafter\def\csname LT7\endcsname{\color{black}}%
      \expandafter\def\csname LT8\endcsname{\color{black}}%
    \fi
  \fi
    \setlength{\unitlength}{0.0500bp}%
    \ifx\gptboxheight\undefined%
      \newlength{\gptboxheight}%
      \newlength{\gptboxwidth}%
      \newsavebox{\gptboxtext}%
    \fi%
    \setlength{\fboxrule}{0.5pt}%
    \setlength{\fboxsep}{1pt}%
    \definecolor{tbcol}{rgb}{1,1,1}%
\begin{picture}(3168.00,2016.00)%
    \gplgaddtomacro\gplbacktext{%
      \csname LTb\endcsname%%
      \put(924,704){\makebox(0,0)[r]{\strut{}\small$0.1$}}%
      \put(924,936){\makebox(0,0)[r]{\strut{}\small$1$}}%
      \put(924,1169){\makebox(0,0)[r]{\strut{}\small$10$}}%
      \put(924,1401){\makebox(0,0)[r]{\strut{}\small$100$}}%
      \put(924,1634){\makebox(0,0)[r]{\strut{}\small$1000$}}%
      \put(924,1796){\makebox(0,0)[r]{\strut{}\small$5000$}}%
      \put(990,484){\makebox(0,0){\strut{}\small 20}}%
      \put(1881,484){\makebox(0,0){\strut{}\small 40}}%
      \put(2771,484){\makebox(0,0){\strut{}\small 60}}%
      \put(322,634){\rotatebox{90}{\makebox(0,0)[l]{\strut{}Time (msec)}}}%
    }%
    \gplgaddtomacro\gplfronttext{%
      \csname LTb\endcsname%%
      \put(1880,154){\makebox(0,0){\strut{}\small Trace length}}%
    }%
    \gplbacktext
    \put(0,0){\includegraphics[width={158.40bp},height={100.80bp}]{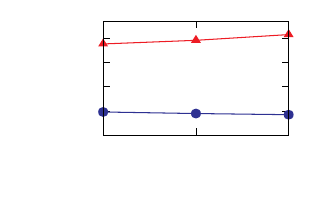}}%
    \gplfronttext
  \end{picture}%
\endgroup

%% file: TIME-LSTM-LENGTH.tex
% GNUPLOT: LaTeX picture with Postscript
\begingroup
  % Encoding inside the plot.  In the header of your document, this encoding
  % should to defined, e.g., by using
  % \usepackage[cp1252,<other encodings>]{inputenc}
  \selectfont
  \makeatletter
  \providecommand\color[2][]{%
    \GenericError{(gnuplot) \space\space\space\@spaces}{%
      Package color not loaded in conjunction with
      terminal option `colourtext'%
    }{See the gnuplot documentation for explanation.%
    }{Either use 'blacktext' in gnuplot or load the package
      color.sty in LaTeX.}%
    \renewcommand\color[2][]{}%
  }%
  \providecommand\includegraphics[2][]{%
    \GenericError{(gnuplot) \space\space\space\@spaces}{%
      Package graphicx or graphics not loaded%
    }{See the gnuplot documentation for explanation.%
    }{The gnuplot epslatex terminal needs graphicx.sty or graphics.sty.}%
    \renewcommand\includegraphics[2][]{}%
  }%
  \providecommand\rotatebox[2]{#2}%
  \@ifundefined{ifGPcolor}{%
    \newif\ifGPcolor
    \GPcolorfalse
  }{}%
  \@ifundefined{ifGPblacktext}{%
    \newif\ifGPblacktext
    \GPblacktexttrue
  }{}%
  % define a \g@addto@macro without @ in the name:
  \let\gplgaddtomacro\g@addto@macro
  % define empty templates for all commands taking text:
  \gdef\gplbacktext{}%
  \gdef\gplfronttext{}%
  \makeatother
  \ifGPblacktext
    % no textcolor at all
    \def\colorrgb#1{}%
    \def\colorgray#1{}%
  \else
    % gray or color?
    \ifGPcolor
      \def\colorrgb#1{\color[rgb]{#1}}%
      \def\colorgray#1{\color[gray]{#1}}%
      \expandafter\def\csname LTw\endcsname{\color{white}}%
      \expandafter\def\csname LTb\endcsname{\color{black}}%
      \expandafter\def\csname LTa\endcsname{\color{black}}%
      \expandafter\def\csname LT0\endcsname{\color[rgb]{1,0,0}}%
      \expandafter\def\csname LT1\endcsname{\color[rgb]{0,1,0}}%
      \expandafter\def\csname LT2\endcsname{\color[rgb]{0,0,1}}%
      \expandafter\def\csname LT3\endcsname{\color[rgb]{1,0,1}}%
      \expandafter\def\csname LT4\endcsname{\color[rgb]{0,1,1}}%
      \expandafter\def\csname LT5\endcsname{\color[rgb]{1,1,0}}%
      \expandafter\def\csname LT6\endcsname{\color[rgb]{0,0,0}}%
      \expandafter\def\csname LT7\endcsname{\color[rgb]{1,0.3,0}}%
      \expandafter\def\csname LT8\endcsname{\color[rgb]{0.5,0.5,0.5}}%
    \else
      % gray
      \def\colorrgb#1{\color{black}}%
      \def\colorgray#1{\color[gray]{#1}}%
      \expandafter\def\csname LTw\endcsname{\color{white}}%
      \expandafter\def\csname LTb\endcsname{\color{black}}%
      \expandafter\def\csname LTa\endcsname{\color{black}}%
      \expandafter\def\csname LT0\endcsname{\color{black}}%
      \expandafter\def\csname LT1\endcsname{\color{black}}%
      \expandafter\def\csname LT2\endcsname{\color{black}}%
      \expandafter\def\csname LT3\endcsname{\color{black}}%
      \expandafter\def\csname LT4\endcsname{\color{black}}%
      \expandafter\def\csname LT5\endcsname{\color{black}}%
      \expandafter\def\csname LT6\endcsname{\color{black}}%
      \expandafter\def\csname LT7\endcsname{\color{black}}%
      \expandafter\def\csname LT8\endcsname{\color{black}}%
    \fi
  \fi
    \setlength{\unitlength}{0.0500bp}%
    \ifx\gptboxheight\undefined%
      \newlength{\gptboxheight}%
      \newlength{\gptboxwidth}%
      \newsavebox{\gptboxtext}%
    \fi%
    \setlength{\fboxrule}{0.5pt}%
    \setlength{\fboxsep}{1pt}%
    \definecolor{tbcol}{rgb}{1,1,1}%
\begin{picture}(3168.00,2016.00)%
    \gplgaddtomacro\gplbacktext{%
      \csname LTb\endcsname%%
      \put(924,704){\makebox(0,0)[r]{\strut{}\small$0.1$}}%
      \put(924,936){\makebox(0,0)[r]{\strut{}\small$1$}}%
      \put(924,1169){\makebox(0,0)[r]{\strut{}\small$10$}}%
      \put(924,1401){\makebox(0,0)[r]{\strut{}\small$100$}}%
      \put(924,1634){\makebox(0,0)[r]{\strut{}\small$1000$}}%
      \put(924,1796){\makebox(0,0)[r]{\strut{}\small$5000$}}%
      \put(990,484){\makebox(0,0){\strut{}\small 20}}%
      \put(1881,484){\makebox(0,0){\strut{}\small 40}}%
      \put(2771,484){\makebox(0,0){\strut{}\small 60}}%
      \put(322,634){\rotatebox{90}{\makebox(0,0)[l]{\strut{}Time (msec)}}}%
    }%
    \gplgaddtomacro\gplfronttext{%
      \csname LTb\endcsname%%
      \put(1880,154){\makebox(0,0){\strut{}\small Trace length}}%
    }%
    \gplbacktext
    \put(0,0){\includegraphics[width={158.40bp},height={100.80bp}]{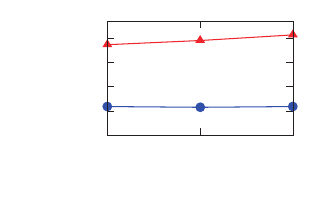}}%
    \gplfronttext
  \end{picture}%
\endgroup

%% file: TIME-ALL-AVGLENGTH.tex
% GNUPLOT: LaTeX picture with Postscript
\begingroup
  % Encoding inside the plot.  In the header of your document, this encoding
  % should to defined, e.g., by using
  % \usepackage[cp1252,<other encodings>]{inputenc}
  \selectfont
  \makeatletter
  \providecommand\color[2][]{%
    \GenericError{(gnuplot) \space\space\space\@spaces}{%
      Package color not loaded in conjunction with
      terminal option `colourtext'%
    }{See the gnuplot documentation for explanation.%
    }{Either use 'blacktext' in gnuplot or load the package
      color.sty in LaTeX.}%
    \renewcommand\color[2][]{}%
  }%
  \providecommand\includegraphics[2][]{%
    \GenericError{(gnuplot) \space\space\space\@spaces}{%
      Package graphicx or graphics not loaded%
    }{See the gnuplot documentation for explanation.%
    }{The gnuplot epslatex terminal needs graphicx.sty or graphics.sty.}%
    \renewcommand\includegraphics[2][]{}%
  }%
  \providecommand\rotatebox[2]{#2}%
  \@ifundefined{ifGPcolor}{%
    \newif\ifGPcolor
    \GPcolorfalse
  }{}%
  \@ifundefined{ifGPblacktext}{%
    \newif\ifGPblacktext
    \GPblacktexttrue
  }{}%
  % define a \g@addto@macro without @ in the name:
  \let\gplgaddtomacro\g@addto@macro
  % define empty templates for all commands taking text:
  \gdef\gplbacktext{}%
  \gdef\gplfronttext{}%
  \makeatother
  \ifGPblacktext
    % no textcolor at all
    \def\colorrgb#1{}%
    \def\colorgray#1{}%
  \else
    % gray or color?
    \ifGPcolor
      \def\colorrgb#1{\color[rgb]{#1}}%
      \def\colorgray#1{\color[gray]{#1}}%
      \expandafter\def\csname LTw\endcsname{\color{white}}%
      \expandafter\def\csname LTb\endcsname{\color{black}}%
      \expandafter\def\csname LTa\endcsname{\color{black}}%
      \expandafter\def\csname LT0\endcsname{\color[rgb]{1,0,0}}%
      \expandafter\def\csname LT1\endcsname{\color[rgb]{0,1,0}}%
      \expandafter\def\csname LT2\endcsname{\color[rgb]{0,0,1}}%
      \expandafter\def\csname LT3\endcsname{\color[rgb]{1,0,1}}%
      \expandafter\def\csname LT4\endcsname{\color[rgb]{0,1,1}}%
      \expandafter\def\csname LT5\endcsname{\color[rgb]{1,1,0}}%
      \expandafter\def\csname LT6\endcsname{\color[rgb]{0,0,0}}%
      \expandafter\def\csname LT7\endcsname{\color[rgb]{1,0.3,0}}%
      \expandafter\def\csname LT8\endcsname{\color[rgb]{0.5,0.5,0.5}}%
    \else
      % gray
      \def\colorrgb#1{\color{black}}%
      \def\colorgray#1{\color[gray]{#1}}%
      \expandafter\def\csname LTw\endcsname{\color{white}}%
      \expandafter\def\csname LTb\endcsname{\color{black}}%
      \expandafter\def\csname LTa\endcsname{\color{black}}%
      \expandafter\def\csname LT0\endcsname{\color{black}}%
      \expandafter\def\csname LT1\endcsname{\color{black}}%
      \expandafter\def\csname LT2\endcsname{\color{black}}%
      \expandafter\def\csname LT3\endcsname{\color{black}}%
      \expandafter\def\csname LT4\endcsname{\color{black}}%
      \expandafter\def\csname LT5\endcsname{\color{black}}%
      \expandafter\def\csname LT6\endcsname{\color{black}}%
      \expandafter\def\csname LT7\endcsname{\color{black}}%
      \expandafter\def\csname LT8\endcsname{\color{black}}%
    \fi
  \fi
    \setlength{\unitlength}{0.0500bp}%
    \ifx\gptboxheight\undefined%
      \newlength{\gptboxheight}%
      \newlength{\gptboxwidth}%
      \newsavebox{\gptboxtext}%
    \fi%
    \setlength{\fboxrule}{0.5pt}%
    \setlength{\fboxsep}{1pt}%
    \definecolor{tbcol}{rgb}{1,1,1}%
\begin{picture}(3168.00,2016.00)%
    \gplgaddtomacro\gplbacktext{%
      \csname LTb\endcsname%%
      \put(858,686){\makebox(0,0)[r]{\strut{}\small$0.1$}}%
      \put(858,875){\makebox(0,0)[r]{\strut{}\small$1$}}%
      \put(858,1065){\makebox(0,0)[r]{\strut{}\small$10$}}%
      \put(858,1254){\makebox(0,0)[r]{\strut{}\small$100$}}%
      \put(858,1444){\makebox(0,0)[r]{\strut{}\small$1000$}}%
      \put(858,1576){\makebox(0,0)[r]{\strut{}\small$5000$}}%
      \put(1287,554){\rotatebox{-45}{\makebox(0,0)[l]{\strut{}MB3}}}%
      \put(1584,554){\rotatebox{-45}{\makebox(0,0)[l]{\strut{}MB5}}}%
      \put(1881,554){\rotatebox{-45}{\makebox(0,0)[l]{\strut{}MB7}}}%
      \put(2177,554){\rotatebox{-45}{\makebox(0,0)[l]{\strut{}MB10}}}%
      \put(2474,554){\rotatebox{-45}{\makebox(0,0)[l]{\strut{}MA}}}%
      \put(248,629){\rotatebox{90}{\makebox(0,0)[l]{\strut{}Time (msec)}}}%
    }%
    \gplgaddtomacro\gplfronttext{%
      \csname LTb\endcsname%%
      \put(994,1843){\makebox(0,0)[r]{\strut{}T}}%
      \csname LTb\endcsname%%
      \put(1849,1843){\makebox(0,0)[r]{\strut{}T+A}}%
      \csname LTb\endcsname%%
      \put(2704,1843){\makebox(0,0)[r]{\strut{}T+R}}%
    }%
    \gplbacktext
    \put(0,0){\includegraphics[width={158.40bp},height={100.80bp}]{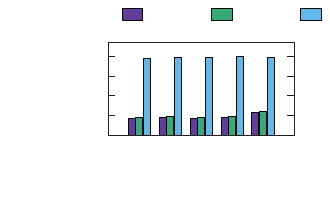}}%
    \gplfronttext
  \end{picture}%
\endgroup